%% file: main.tex
\documentclass[10pt,twocolumn,letterpaper]{article}

\usepackage{cvpr}              

\input{preamble}

\definecolor{cvprblue}{rgb}{0.21,0.49,0.74}
\usepackage[pagebackref,breaklinks,colorlinks,citecolor=cvprblue]{hyperref}

\usepackage[accsupp]{axessibility}
\usepackage{graphicx}
\usepackage{amsmath}
\usepackage{amssymb}
\usepackage{booktabs}
\usepackage{xcolor}
\usepackage{pifont}
\usepackage{multirow}
\usepackage{float}
\usepackage{array}
\usepackage{tikz}
\usepackage[thinc]{esdiff}
\newcommand{\nrange}[4][]{%
  #2=%
  \ifblank{#1}{
    #3, \dots ,#4%
  }{%
    #3, #1, \dots,#4%
  }%
}

\def\paperID{10152} 
\def\confName{CVPR}
\def\confYear{2024}

\title{Your Image is My Video: Reshaping the Receptive Field via Image-To-Video
Differentiable AutoAugmentation and Fusion}

\author{Sofia Casarin$^1$,
Cynthia I. Ugwu$^1$,
Sergio Escalera$^{2,3}$,
Oswald Lanz$^1$ \\
$^1$Free University of Bozen-Bolzano, Bolzano, Italy \\
$^2$Computer Vision Center, Barcelona, Spain \\
$^3$Universitat de Barcelona, Barcelona, Spain \\
{\tt\small \{scasarin, cugwu, olanz\}@unibz.it, sergio@maia.ub.es}
\and
}

\begin{document}
\maketitle
\input{sec/0_abstract}    
\input{sec/1_intro}
\input{sec/2_RL}
\input{sec/3_method}

\input{sec/4_experiments}

\input{sec/Conclusions}

{
    \small
    \bibliographystyle{ieeenat_fullname}
    \bibliography{main}
}
\input{sec/X_suppl}


\end{document}

%% file: preamble.tex
\usepackage[dvipsnames]{xcolor}


%% file: sec/0_abstract.tex
\begin{abstract}
The landscape of deep learning research is moving towards innovative strategies to harness the true potential of data. Traditionally, 
emphasis has been on scaling model architectures, resulting in large and complex neural networks, which can be difficult to train with limited computational resources. However, independently of the model size, data quality (\emph{\ie} amount and variability) is still a major factor that affects model generalization.
In this work, we propose a novel technique to exploit available data through the use of automatic data augmentation for the tasks of image classification and semantic segmentation. 
We introduce the first Differentiable Augmentation Search method (DAS) to generate variations of images that can be processed as videos. Compared to previous approaches, DAS is extremely fast and flexible, allowing the search on very large search spaces in less than a GPU day.
Our intuition is that the increased receptive field in the temporal dimension provided by DAS could lead to benefits also to the spatial receptive field. 
More specifically, we leverage DAS to guide the reshaping of the spatial receptive field by selecting task-dependant transformations.
As a result, compared to standard augmentation alternatives, we improve in terms of accuracy on ImageNet, Cifar10, Cifar100, Tiny-ImageNet, Pascal-VOC-2012 and CityScapes datasets when plugging-in our DAS over different light-weight video backbones. 
\vspace{-0.3cm}
\end{abstract}

%% file: sec/1_intro.tex
\section{Introduction}
\label{sec:intro}

Creating models with significantly increased capacity, in an attempt to achieve incremental performance improvements, has been the prevailing approach in designing Convolutional Neural Network (CNN) classifiers. 
As a result, CNNs with increased depth~\cite{szegedy2016rethinking, he2016deep, simonyan2014very, Zhang_2022}, 
and Vision Transformers (ViTs)~\cite{dosovitskiy2020image} were proposed over the years. Specifically, ViT has demonstrated promising results on a wide variety of computer vision tasks including image classification and semantic segmentation 
\cite{touvron2021going, dosovitskiy2020image, oquab2023dinov2, wortsman2022model}. \noindent
\begin{figure}[t]
  \centering
  \begin{subfigure}{\linewidth}
    \includegraphics[width=1\linewidth]{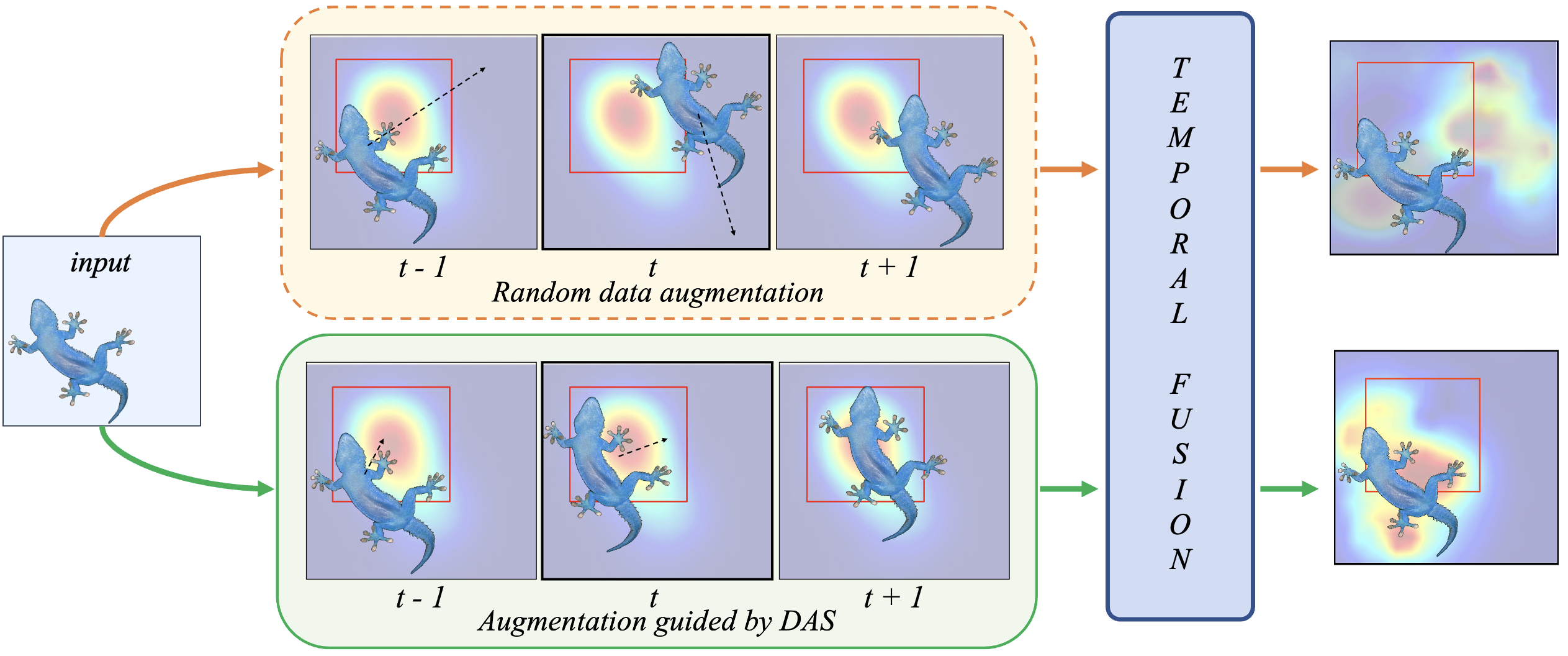}
    \caption{Conceptual representation of the proposed approach. We reshape the Receptive Field (RF) by applying affine transformations optimized through our Differentiable Augmentation Search (DAS). On the top right you can see how fusing with random transformation would not lead to benefits as, when concatenating in time, the employed shift mechanism would fuse features related to random parts. On the bottom, the augmentations guided by DAS obtain specific shapes of the RF so that more context is kept.}
     \label{fig:abs_fig}
  \end{subfigure}
  \\

  \begin{subfigure}{\linewidth}
  \begin{subfigure}{0.24\linewidth}
    \includegraphics[width=\linewidth,trim={50 50 50 50},clip]{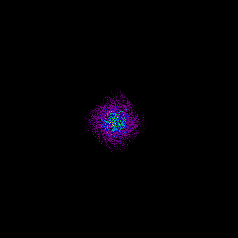}
  \end{subfigure} 
  \begin{subfigure}{0.24\linewidth}
    \includegraphics[width=\linewidth]{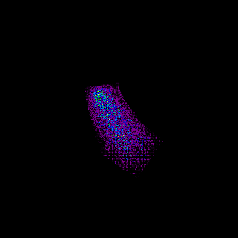}
  \end{subfigure} 
  \begin{subfigure}{0.24\linewidth}
    \includegraphics[width=\linewidth]{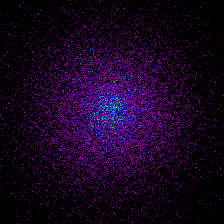}
  \end{subfigure} 
    \begin{subfigure}{0.24\linewidth}
    \includegraphics[width=\linewidth]{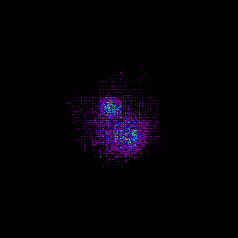}
    \put(-217,3){\color{white}\footnotesize rotate}
    \put(-161,3){\color{white}\footnotesize translate} 
    \put(-96,3){\color{white}\footnotesize zoom}
    \put(-51,3){\color{white}\footnotesize found by DAS}
  \end{subfigure}
  \caption{RF visualization (ResNet-50, with GSF fusion) when different single or composed transformations are applied. The last column shows our DAS selected operation for CIFAR-10 and CIFAR-100, which combines translation, rotation and zoom. More details in~\cref{tab:c10_c100_tiny}. }
  \label{subfig:RF_shapes}
   \end{subfigure}
  \caption{\ref{fig:abs_fig}~overviews our approach and \ref{subfig:RF_shapes}~shows a real example of obtained receptive fields. The employed transformations are fundamental to shape the receptive field, as shown in~\ref{subfig:RF_shapes}. The augmented images with DAS (Sec.~3.1) are concatenated in time, and processed through a video network that partially shifts and fuses the features (Sec.~3.2).}\vspace{-0.3cm}
  \label{fig:empirical_RF}
\end{figure}
However, while these techniques have demonstrated remarkable success, it is noteworthy that these high-capacity models necessitate increased computational resources for effective training and inference, making them economically impractical for training and deployment within practical application scenarios. Moreover, the over-parametrization of ViT and Deep CNNs makes the networks prone to overfitting, thus requiring strong regularization 
to achieve ideal performance. As a result, one alternative trend relies in exploiting the power of data. In this work, we tackle this problem and propose a novel way of augmenting data with the goal of expanding the receptive field of CNNs. \\ \noindent
Data augmentation techniques, usually employed to enhance the generalization capabilities of machine learning models, rely on meticulous design, necessitating domain-specific knowledge. To this aim, various auto data augmentation methods, inspired by Neural Architecture Search (NAS), have been extensively utilized in the field of supervised learning~\cite{cubuk2020randaugment, cubuk2019autoaugment, muller2021trivialaugment, lingchen2020uniformaugment} to search for accurate augmentation strategies.  However, many of these approaches, either require much time~\cite{cubuk2019autoaugment, ho2019population, tian2020improving}, forcing the search to be performed on a proxy task or do not propose an effective search strategy, but rather a well defined search space as in~\cite{cubuk2020randaugment}. The former, makes a strong assumption that the proxy task provides a predictive indication of the larger task~\cite{cubuk2020randaugment}. The dramatic reduction in parameter space of the latter, which allows simple grid search, implies a strong knowledge in the search space definition. As a result, as we empirically show, including ``noisy" transformations would imply much larger searching time to achieve the same performance.
By harnessing affine transformations optimized through our new Differentiable Augmentation Search (DAS) strategy, we generate a series of images that exhibit motion within a specific region, treating them analogously to video sequences. Inspired by transformation-based models~\cite{Joost_2017} that, operating in the space of affine transforms, re-phrase the frames prediction problem as modelling transformation within frames, we introduce a novel methodology to extend the Receptive Field (\cref{fig:abs_fig}).  With the underlying hypothesis that augmenting the temporal dimension's receptive field (RF) could potentially yield advantages for the spatial RF as well, we establish the viability of this approach and empirically showcase its merits by employing Video Networks for processing such image sequences. In this context, to alleviate the unnecessary increase of computational overhead for image-classification and semantic segmentation tasks, we exploit a feature shift mechanism~\cite{sudhakaran2022gate}, which in the context of video action recognition tasks demonstrates comparable performance to a 3D CNN while keeping 2D CNN complexity.
In summary, our contributions are: 
\begin{itemize}
    \item  We re-formulate the automatic data augmentation field in a differentiable manner, and propose DAS. By defining a continuous search space of image transformations and exploiting a perturbation-based approach for the transformation selection, we provide a very general and easy-to-deploy alternative to slower existing reinforcement learning methods, and to more search space definition sensitive random ones. 
    \item We propose a new way of handling 2D data by repeatedly transforming and concatenating images as frames in a video, obtaining a new perspective to exploit the richness of data. To this aim, we address the question of how the increase of the receptive field in a third dimension impacts the original 2D spatial receptive field.
    \item We successfully expand, as a result, the receptive field for image classification and segmentation tasks. This allows obtaining ResNet-152 state-of-the-art results for ImageNet while employing a ResNet-50 temporal expanded network, having less than half-parameters, and surpassing DeepLabv3 by 1.3 \% on Pascal-VOC and ResNest models by 1.1\% on CityScapes datasets.
\end{itemize}

%% file: sec/2_RL.tex
\section{Related Work}
\label{sec:related_works}
In our work, we are interested in 
image classification and semantic image segmentation tasks. A large focus of the computer vision community has been on engineering better network architectures to improve performance. In earlier designs ImageNet progress was dominated by CNNs~\cite{krizhevsky2012imagenet, simonyan2014very, szegedy2015going, jian2016deep, tan2019efficientnet, howard2017mobilenets} and more prominently now Vision Transformers ~\cite{dosovitskiy2020image, swin} have reached competitive results 
due to their wider receptive fields. 
\noindent
Semantic segmentation requires either global features or contextual interactions to accurately classify, at the pixel level, objects at multiple scales. To this aim, atrous convolutions~\cite{chen2017rethinking, chen2018encoder}, spatial pyramid pooling modules~\cite{zhao2017pyramid}, and networks with attention modules~\cite{Zhang_2022, fu2019dual, wang2211internimage} with almost 1 billion parameters were proposed. While these methods enhance the receptive field, achieving good results in terms of performance, they are extremely data hungry and often require strong regularization techniques, such as data augmentation. 

\begin{figure*}
  \centering
    \includegraphics[width=1\linewidth]{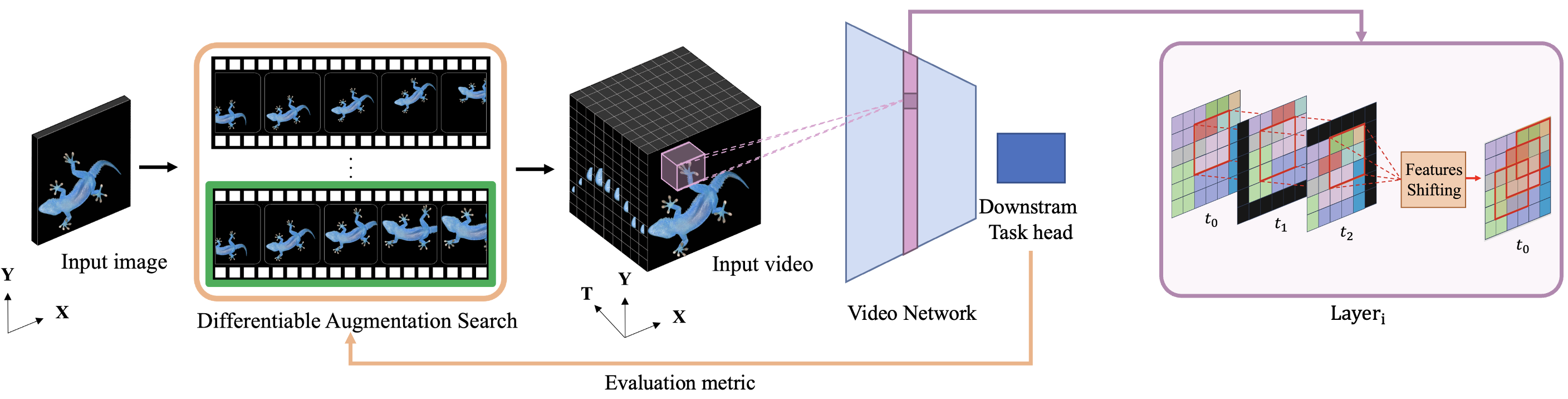}
    \caption{Our method takes an input image and processes it through a DAS cell. The cell, as shown more in detail in Fig.~\ref{fig:das}, applies all possible transformations defined in the search space and generates an input video. The video is processed through a video network integrated by a temporal shift mechanism, with the goal of shifting the features of adjacent frames. As it can be observed in the pink box, the features shifted and combined as if a kernel $3\times3\times3$ was applied. As the content derives from transformations of the same image, the result over the original 2D image is a reshaping of the RF. Finally, the predictions for the video input are combined so that the performance for the original 2D task are given back as feedback to the DAS cell.}
    \label{fig:method}
  \vspace{-0.5cm}
\end{figure*}

\subsection{Automatic Data Augmentation}
Traditionally, data augmentation requires manual design and domain knowledge. Random cropping, image mirroring, and color distortion are common in natural image datasets like CIFAR-10 and ImageNet. Elastic distortion and affine transformations such as translation and rotation are more common in datasets like MNIST\cite{lecun1998gradient} and SVHN~\cite{netzer2011reading}.
Many methods have been influenced by NAS~\cite{zoph2016neural} to find the best dataset-specifc set of augmentation policies/strategies ~\cite{cubuk2019autoaugment, lim2019fast, tian2020improving, lin2019online, ho2019population}. AutoAugment (AA)~\cite{cubuk2019autoaugment}, the first automatic augmentation method, uses reinforcement learning to predict accurate problem-dependant augmentation policies. Despite its success, the approach has the main drawback of running the search on a smaller version of the datasets due to the multi-level search space and the repetitive training, making a strong assumption that the proxy task provides a predictive indication of the larger task. Methods like~\cite{lingchen2020uniformaugment, cubuk2020randaugment, muller2021trivialaugment} drastically reduce the parameter space for data augmentation, which allows the methods to be trained on the full dataset. In RandAugmnet (RA)~\cite{cubuk2020randaugment} only two hyperparameters are used, one controlling the number of augmentations to combine for each image, and the other controlling the magnitude of operations. The downside of RA is that it performs a grid search over a set of augmentation operations incurring up to $\times80$ overhead over a single training~\cite{muller2021trivialaugment}. 
UniformAugment (UA)~\cite{lingchen2020uniformaugment} and TrivialAugment (TA)~\cite{muller2021trivialaugment}, instead, propose parameter-free algorithms where hyperparameters are sampled uniformly in the augmentation space.
Similarly to UA, we define a continuous augmentations space. However, differently from UA, RA and AA, we do not randomly sample the \emph{ad-hoc} hyperparameters or perform the search on a reduced dataset, but use a continuous search strategy which makes our method fast without the need to reduce drastically the search space or to include user designer bias.
\subsection{Enhancing the Receptive Field}
In the context of CNNs, the general trend that led to very deep networks is motivated by the seek for a receptive field expansion.
The concept of RF is indeed important for understanding and diagnosing how deep CNNs work as it determines what information a single neuron has access to. Over the years, many attempts were made to expand such a field, either by increasing the factors determining its theoretical size, \eg the depth of a neural network and the kernel size, or by providing more context information.\vspace{-0.4cm}
\paragraph{Spatial Domain}
In the context of image classification and semantic segmentation, using context information from the whole image can significantly help improve the performance. Mostajabi \etal~\cite{Mostajabi_2014} demonstrated that by using the ``zoom-out" features they can achieve impressive performance for the semantic segmentation task. Liu \etal~\cite{Liu2015ParseNetLW} noticed that, although theoretically, the features from the top layers of a neural network should have very large receptive fields, in practice the empirical size is much smaller and consequently not enough to capture the global context. To this aim, they propose to use global averaging 
to pool the context features from layers of the neural network, proving empirically that such a method results in a larger empirical RF. In~\cite{Richter_2021}, Richter \etal question the actual need for a very deep network and propose a method for layer pruning based on the analysis of the size of the RF. \vspace{-0.4cm}
\paragraph{Temporal Domain}
How to effectively expand the receptive field in the temporal domain has been investigated in video understanding research
~\cite{Wang_2016, Luo_2019, Lin_2019, Sudhakaran_2020, sudhakaran2022gate}. 3D CNNs can capture three-dimensional features, however, they are more computationally intensive than 2D CNN\'s
video networks. In~\cite{Lin_2019} the authors implement the Temporal Shift Module (TSM) where a 2D CNN is used as a backbone to extract spatial information. The temporal features are integrated by shifting a fixed amount of channels forward and backward along the temporal dimension. Subsequently, 
Sudhakaran \etal~\cite{sudhakaran2022gate} proposed the Gate-Shift-Fuse (GSF) module, a spatiotemporal feature extraction that leverages on a learnable shift of the channels on the temporal dimension as well as channel weighting to fuse the shifted features. TSM and GSF are lightweight and therefore good video network candidates for our image-to-video pipeline in~\cref{fig:method}.\vspace{0.2cm}

In contrast to prior methods that enhance the spatial receptive field by concatenating global or zoomed-out features or that utilize deeper networks, 
our method extends beyond by aggregating features derived from diverse transformations and by processing them 
as a video. We employ distinct transformations to generate variations of the input. Given that data augmentation necessitates domain-specific knowledge, we adopt automatic data augmentation. 

%% file: sec/3_method.tex
\section{Methods}
\label{sec:method}

In this section, we first introduce our new differentiable transformation search process effectively responsible for the reshape of the Receptive Field for image classification and image segmentation problems (\cref{sec:DAS_method}). We highlight the benefit of a continuous search space and motivate the importance of a perturbation-based selection technique. We then detail in~\cref{sec:TDA} how we make use of DAS to exploit the power of data, showing that with respect to traditional methods expanding the RF, our effect results in a ``reshaping" rather than an expansion. The pipeline of the proposed architecture is shown in~\cref{fig:method}.

\subsection{Differentiable Augmentation Search}\label{sec:DAS_method}
\begin{figure}
  \centering
   \includegraphics[width=\linewidth]{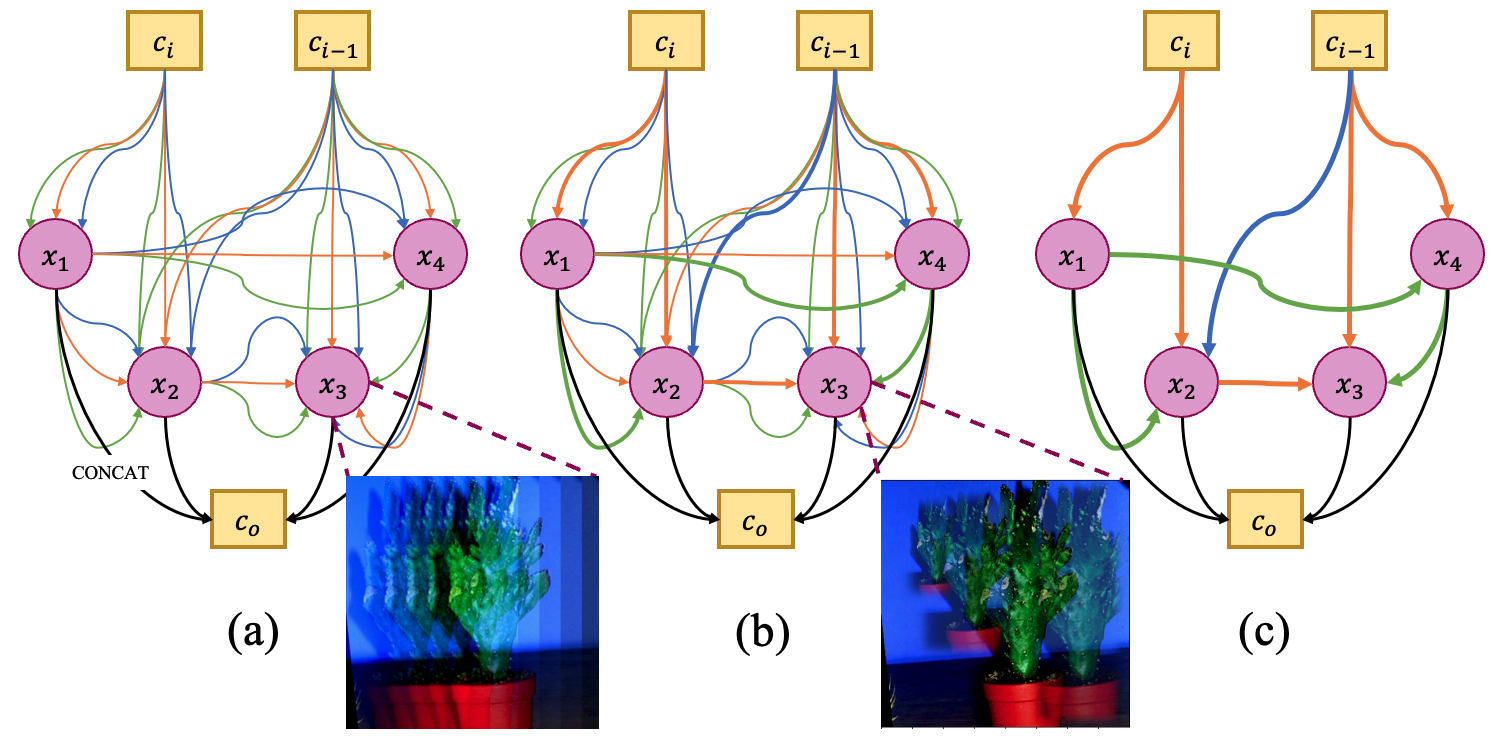}
  \caption{Cell structure in DAS. 
  Multiple operations are defined on each edge, collectively applied to the image, and optimized during training: as the gradients are updated through multiple steps, the $\tau$ values associated with each operation change. Fig. {\color{red}3b} depicts one step of such process through thicker edges. In the end, the cell is discretized through a perturbation-based approach, the final operations are chosen and composed (black edges).}
  \label{fig:das}
\end{figure}

In order to extend images to video and to properly reshape the RF (Fig.~\ref{fig:abs_fig}), a set of optimal transformations needs to be found. 
We propose a new general and differentiable approach, later deployed with a restricted search space for our use case, searching for the transformations generating the ``best video" to process for a given video network and downstream task. 
Inspired by Differentiable Neural Architecture Search~\cite{liu2018darts, wang2021rethinking} we define a continuous search space of transformations, which leads to a differentiable learning objective for the joint optimization of the transformations to be applied and the weights of the architecture. Following~\cite{liu2018darts, zoph2018learning}, we search for a computation cell as the initial block that generates the input for the chosen video network. \\
A cell, depicted in Fig.~\ref{fig:das}, is a directed acyclic graph consisting of an ordered sequence of $N$ nodes. Each node $x^{(i)}$ in the cell is the transformed image, and each directed edge $(i,j)$ is associated with a data augmentation technique. For the deployment of our Differentiable Architecture Search, we defined two search spaces. The first one, similarly to AA and RA methods~\cite{cubuk2019autoaugment, cubuk2020randaugment}, comprises as set of augmentations Shear X/Y, Translate X/Y, Rotate, AutoContrast, Invert, Equalize, Solarize, Posterize, Color, Brightness, Sharpness, Cutout, and Identity that corresponds to applying no transformation. The cell has two input nodes and a single output node. The size of such a search space is $13^{14}$. The second search space, deployed for our purpose, includes Translate X/Y, Scale, Rotate, and Identity. It employs a cell with one single input and output node, having a size of $5^{10}$. Given the set $\mathcal{T}$ of  candidate transformations, the categorical choice of applying a transformation is relaxed to a Softmax of all possible transformations $t \in \mathcal{T}$. For a given edge $(i,j)$, the transformation to be applied to the input $x$ is expressed as:
\begin{align}
    \bar{t}^{(i, j)}(x) &= \sum_{t \in \mathcal{T}} \frac{\exp(\tau_{t}^{(i ,j)})}{\sum_{t \in \mathcal{T}} \exp(\tau_{t'}^{(i,j)})} \cdot t(x),
\end{align}
where the weights of a transformation are parameterized by a vector $\tau^{(i,j)}$ of dimension $|\mathcal{T}|$. The famous bi-level optimization problem~\cite{colson2007overview} of NAS is re-formularized as: 
\begin{equation}
    \begin{gathered}
        \min_{\boldsymbol{\tau}} \quad \mathcal{L}_{\text{val}}(\boldsymbol{\tau}, \boldsymbol{w}^*(\boldsymbol{\tau})) \\
        \text{s.t.} \quad \boldsymbol{w}^*(\boldsymbol{\tau}) = \arg\min_{\boldsymbol{w}} \mathcal{L}_{\text{train}}(\boldsymbol{w}, \boldsymbol{\tau}),
    \end{gathered}
\end{equation}

where the model weights $\boldsymbol{w}$ and the transformations parameters $\boldsymbol{\tau}$ are jointly optimized via gradient updates following standard DARTS procedure. We would like to stress that, differently from the original DARTS, here there is no real architecture optimization. The real CNN network that processes the data is fixed and chosen a priori. What we are looking for is a cell where only transformations are applied, and we treat this cell as a sort of ``stem" block, placed before the backbone.
At the end of the search phase, the best transformations are not chosen by selecting the largest $\tau$ value, as this was shown in~\cite{wang2021rethinking} to be based on a wrong assumption, \ie the $\tau$ representing the strength of a transformation agrees with the discretization accuracy at convergence. 
We rather deploy a perturbation-based approach, where the transformation importance is evaluated in terms of its contribution to the neural network performance. To this aim, for each transformation on a given edge, we mask it while preserving all other transformations, then re-evaluate the cell+CNN. The operation resulting in the greatest reduction in network validation accuracy is identified as the pivotal operation on that edge. Let us indeed consider a cell from a simplified search space composed of only two transformations:
\begin{equation}
   \begin{aligned}
\begin{array}{c}
  \mathbf{I} = \begin{bmatrix}
    1 & 0 & 0 \\
    0 & 1 & 0 \\
    0 & 0 & 1 \\
  \end{bmatrix}
\end{array}
&
\begin{array}{c}
  \mathbf{T} = \begin{bmatrix}
    1 & 0 & t_x \\
    0 & 1 & t_y \\
    0 & 0 & 1 \\
  \end{bmatrix}
\end{array}
\end{aligned} 
\end{equation}
where $x_I$ represents the output of $I\cdot x$, and $x_T$ the translated output $T \cdot x$. Assume $m^*$ to be the optimal feature map, shared across all edges according to the unrolled estimation view. The current estimation of $m^*$ can then be written as:
\begin{equation}
    \overline{m}(x) = \frac{e^{\tau_T}}{e^{\tau_T} + e^{\tau_I}}x_T + \frac{e^{\tau_I}}{e^{\tau_T} + e^{\tau_I}}x_I
    \label{eq:m_bar}
\end{equation}
The optimal ${\tau^*_T}$ and ${\tau^*_I}$ minimizing the var$(\overline{m}(x) - m^*)$ variance between the optimal feature map $m^*$ and the current estimation $\overline{m}(x)$ are:
\begin{align}
    {\tau^*_I} &\propto \text{var}(x_T - m^*) \label{eq:alpha_I} \\
    {\tau^*_T} &\propto \text{var}(x_I - m^*) \label{eq:alpha_T}
\end{align}
We refer to the Supplementary material for a detailed proof. As in the original paper, also for the transformation search space it holds that, from~\cref{eq:alpha_I} and~\cref{eq:alpha_T}, we can see that the relative magnitudes of $\tau_I$ and $\tau_T$ come down to which one of $x_I$ or $x_T$ is closer to $m^*$ in variance. As $x_I$ comes from the mixed output of a previous edge, and the goal of every edge is to estimate $m^*$ (unrolled estimation) $x_I$ is also directly estimating $m^*$. $x_T$ is the output of a single transformation instead of the complete mixed output of edge $e$, so even at convergence it will deviate from $m^*$. Therefore, if we choose the largest $\tau$ as indicating the best transformation, the algorithm will naturally be led to choose identities. Therefore, as on one hand, including the identity in the search space is fundamental to remove the a-priori bias that a transformation is always needed, on the other hand, as we show in Sec.~\ref{sec:ablation} a cell consisting of only identity transformations leads to poor performance, motivating our transformation selection choice.

\subsection{Temporal Data Augmentation}\label{sec:TDA}
Our goal is to expand and reshape the spatial receptive field. To this aim, we generate videos by  picking from a set of transforms, summarized in our search space defined by  Translate X/Y, Scale, Rotate, and Identity. We included in our search-space commonly used transformations to model motion in videos~\cite{Joost_2017}. We added scaling, motivated by the general difficulty of segmenting objects at different scale, and identity, to remove the bias that the image \emph{always} benefits from applying a transformation. \\ \noindent
Each transformation is applied $\times T$ times, a hyperparameter that defines the length of the generated video. For example, given an image $I(x,y)$, a frame $V(t, x, y)$ of a video applying a vertical translation $\delta y$ is obtained as:
\begin{equation}
    V(t, x, y) = I(x, y - t\delta y)\quad  \nrange{t}{1}{T}.\label{eq:video}
\end{equation}

\begin{figure}[t]
  \centering
   \includegraphics[width=\linewidth]{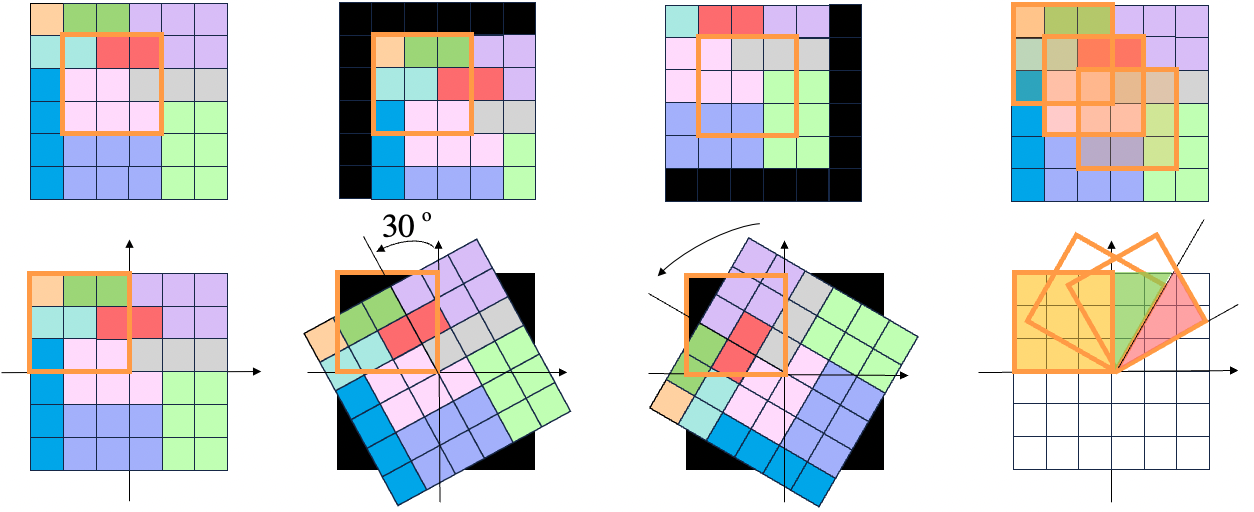}
       \vspace{-0.7cm}
   \caption{RF reshape for translation (top-row) and rotation (bottom-row). In our representation we are assuming that GSF was inserted once, therefore the temporal RF is expanded by 2 and three adjacent frames are considered. The first column shows the features of original image ($f_0$), while second and third columns show the features of frames obtained applying the transformation. The last column shows the effect of processing the input with a temporal shift mechanism.}
   \label{fig:RF}
       \vspace{-0.3cm}
\end{figure}

As Fig.~\ref{fig:method} highlights, the video is given as input to a video network that extracts the features and produces a prediction for the original classification/segmentation task task. For the image classification task, we average the predictions for each frame, while for the semantic segmentation task we first ``undo" the transformation to preserve the locality concept. Although a general video network can process the video input, as we show in our experiments, to reduce the complexity and keep efficiency in our tasks of interest, we deploy a 2D backbone with a temporal-shift mechanism, \ie GSF, integrated. Such a technique is well-established in the domain of video understanding, allowing to achieve the performance of 3D CNN but maintaining 2D CNN's complexity. 
As the authors of~\cite{Lin_2019, sudhakaran2022gate} claim, for each inserted GSF, the temporal RF will be enlarged by 2 as if running a convolution with the kernel size of 3 along the temporal dimension. Therefore, part of the features among 3 adjacent frames is mixed. Let us now focus on what the content of those frames is. Fig.~\ref{fig:RF} gives a proof of concept of our claim for the operations included in our search space. 
As the content in the adjacent frames is nothing but the same image either translated, scaled, or rotated, the increase in size in the temporal RF can be mapped in an augmentation in size and reshape of the spatial RF. 
Let us consider the case of a single-path network, for simplicity. The theoretical RF~\cite{araujo2019computing} size can be expressed as depending only on the stride $s$ and kernel size $k$: 
\vspace{-0.3cm}
\begin{equation}
\begin{split}
    r_{0} &= \sum_{l=1}^{L}((k_i - 1) \prod_{j=1}^{l-1} {s_j}) + 1 \\
\end{split}\vspace{-0.2cm}
\label{eq:u0v0}
\end{equation}
Let us now consider the situation where one GSF module has been inserted after a 2D convolution with kernel $3\times3$, stride $s=1$. Assuming we are considering the first convolutional layer, the theoretical RF for each frame will have a size of 3. Therefore $r_{1}^{f0} = r_{1}^{f1} = r_{1}^{f2} = 3$. As Fig.~\ref{fig:RF} shows, the same region now covers different features, that depend on the applied transformation. As these features are mixed through the GSF mechanism, if we map back to the space of frame $f_0$, the spatial receptive field will have a size equal to $RF_{1} = 3\times r_{1}^{f0} - \bigcap_{i=0}^{2}A(r_{1}^{fi})$, where $ \bigcap_{i=0}^{2}A(r_{1}^{fi})$ denotes the intersecting area among the receptive field of the different frames. If we define as $L = (x_1,y_1)$ the bottom left corner, and $R = (x_2, y_2)$ the top right corner, the intersection area, for the translation case, between two RFs can be simply found as:
\begin{equation}
\begin{split}
    A = (\min(x_1, x_1{'}) - \max(y_1, y_1{'}))\times\\(\min(x_2, x_2{'}) - \max(y_2, y_2{'}))\end{split}
\label{eq:interesection}
\end{equation}
where for the case of a translation of $t_{x}$ along the x-axis and $t_{y}$ over the y-axis, $x_{i}{'} = x + t_x$,  $y_{i}{'} = x + t_y$. For a rotation of $\theta$ degrees counter clockwise, around the point (0,0) (as depicted in~\ref{fig:RF}) $x_{i}{'} = x\cos(-\theta) - y\sin(-\theta)$,  $y_{i}{'} = x\sin(-\theta) + y\cos(-\theta)$. The intersection area calculation however cannot be generalized in this case, as the shape of the intersection polygon depends on the rotation angle. Finally, for the scaling operation the $RF_{1} = \gamma^2 \times (r_{1}^{f0})\times\gamma^2 \times (r_{1}^{f0})$. Please, note that as some components of the RFs overlap (particularly visible for the case of scaling), what is happening is a reshape of the RF, rather than a re-size, as some parts will contribute more than others. Indeed, we need to recall that there is a difference between the theoretical RF and the \emph{Empirical Receptive Field} ERF, defined as $\frac{\partial y(0,0,0)}{\partial x^0(i,j,z)}$, \ie how much $y(0,0,0)$ changes as $x^0(i,j,k)$ changes by a small amount. Here, (i, j, k) is the voxel on a pth layer, y is the output and $x^0$ is the input layer.
If we consider the previous example of a 2D convolution with $k=3\times3$, $s=1$, $t_x = t_y=1, \theta=30^\circ, \gamma=2$: the new receptive field size for the translation is $RF_{1} = 19$, for the rotation is $RF_{1}=14,19$, and for the scaling is $RF_{1} = 144$. As we show in~\cref{sec:ablation}, our method reshapes the RF at the cost of a negligible increase \# of parameters with respect to standard 2D backbones processing images. However, in terms of memory occupation a trade-off is necessary. We empirically find that expanding the image $\times 5$ times well balances accuracy and memory-occupation.\noindent
To sum up, we propose a method to expand the spatial receptive field using data augmentation techniques to generate videos. To this aim, we propose a new auto-augmentation method, namely DAS, that, to the best of our knowledge, is the first differentiable auto-augmentation approach proposed in the field of image classification. The ``fake videos" are processed by a 2D backbone with an integrated temporal shift mechanism, achieving high performance in video-understanding tasks while keeping 2D CNNs complexity. We finally demonstrate that the increased temporal RF from video processing corresponds to an augmented spatial RF, with the extent determined by the specific transformation used. As we will show in~\cref{sec:exp}, such a procedure results in lightweight models that reach state-of-the-art and allow reducing the number of parameters typically increased to expand the receptive field, such as kernel size and depth of a network.

%% file: sec/4_experiments.tex
\section{Experiments}\label{sec:exp}
In this section, we investigate the performance of our temporal expansion approach and our automatic augmentation approach for two visual tasks: image classification and semantic segmentation. Best results are highlited in bold in the tables. 
We also validate our approach through a series of ablation studies. We refer to the Supplementary materials for all the implementation details, more qualitative results, and for additional experiments on the robustness of DAS.

\begin{table}[t]
  \centering\scriptsize
  \begin{tabular}{ l >{\centering\arraybackslash}m{1.2cm} >{\centering\arraybackslash}m{1.2cm}>{\centering\arraybackslash}m{1.2cm}}
    \specialrule{1.1pt}{0pt}{0pt}
    Method & \# Params & FLOPs & Top-1 Acc. \\
    \specialrule{1.1pt}{0pt}{0pt}
    ResNet-50~\cite{he2016deep} & 25.6 & 4.11G & 76.30 \\
    SE-ResNet-50~\cite{hu2018squeeze} & 28.1 & - & 76.90 \\
    Inception-v3~\cite{szegedy2016rethinking} & 27.2 & 11.46G & 77.12 \\
    BnInception~\cite{ioffe2015batch} & - & - & 77.41 \\
    Oct-ResNet-50~\cite{chen2019drop} & 25.6 & - & 77.30 \\
    ResNeXT-50 (32$\times$4d)~\cite{xie2017aggregated} & 25.0 & 8.52G & 77.80 \\
    Res2Net-50 (14w$\times$8s)~\cite{gao2019res2net} & - & - & 78.10 \\
    \specialrule{0.1pt}{0pt}{0pt}
    ResNet-101~\cite{he2016deep} & 44.6 & 7.86G & 77.40 \\
    ResNet-152~\cite{he2016deep} & 60.2 & 11.60G & 78.30 \\
    SE-ResNet-152~\cite{hu2018squeeze} & 67.2 & - & 78.40 \\
    ResNeXt-101 (32$\times$4d)~\cite{xie2017aggregated} & 88.8 & 32.95G & 78.80 \\
    AttentionNeXt-56~\cite{wang2017residual} & 31.9 & - & 78.8 \\
    \specialrule{0.1pt}{0pt}{0pt}
    ViT-L-32~\cite{dosovitskiy2020image} & 304 & 61.55G & 79.66 \\
    \specialrule{0.1pt}{0pt}{0pt}
    FFC-ResNet-50~\cite{chi2020fast} & 26.7 & 5.49G & 77.80 \\
    FFC-ResNeXt-50~\cite{chi2020fast} & 28.0 & 5.66G & 78.00 \\
    FFC-ResNet-101~\cite{chi2020fast} & 46.1 & 9.23G & 78.80 \\
    FFC-ResNet-152~\cite{chi2020fast} & 62.6 & 12.96G & 78.90 \\
    \specialrule{0.1pt}{0pt}{0pt}
    (Ours) - BnInception & - & - & 78.12\\
    (Ours)- Inception-v3 & 27.3 & 11.51G & 78.66\\
    (Ours) - ResNet-50 & 25.7 & 4.2G &79.45\\
    (Ours) - ResNet-101 & 44.4 & 7.96G & 80.05\\
    \textbf{(Ours) - ResNet-152} & 60.4 & 11.66G & \textbf{80.13} \\

    \specialrule{1.1pt}{0pt}{0pt}
    
  \end{tabular}
  \vspace{-0.2cm}
  \caption{Plugging our method into state-of-the-art networks on ImageNet. The first two
sets are top-1 accuracy scores obtained by various state-of-the-art methods, which we transcribe from the
corresponding papers. Deeper models are listed in the second set. The third set reports the performances of
plugging~\cite{chi2020fast}, and last set shows the effect of employing DAS+2D backbone+GSF.}
  \label{tab:ImageNet}
\end{table}

\begin{figure}[t]
  \centering
   \includegraphics[width=7cm,height=3.5cm]{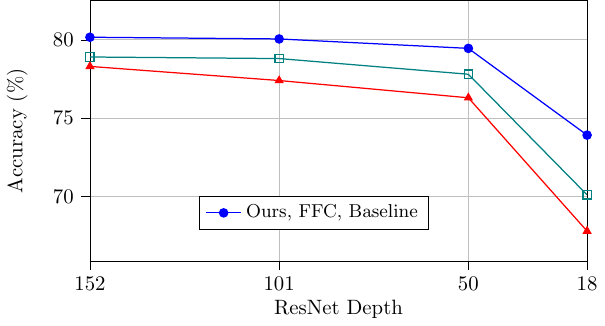}
  \vspace{-0.3cm}
   \caption{ImageNet results for different Resnet depths.}
   \label{fig:depth_ImageNet}
       \vspace{-0.1cm}
\end{figure}

\subsection{Comparison with SOTAs}\label{sec:exp_sota}

\begin{table}[t]
  \centering\small
  \begin{tabular}{ l >{\centering\arraybackslash}m{1.2cm} >{\centering\arraybackslash}m{1.2cm}}
    \specialrule{1.1pt}{0pt}{0pt}
    Method  & FLOPs & mIOU \\
    \specialrule{1.1pt}{0pt}{0pt}
    Adelaide\_VeryDeep\_FCN\_VOC~\cite{wu2016bridging} & - & 79.10 \\
    DeepLabv2-CRF~\cite{Chen_2018} & - & 79.70 \\
    CentraleSupelec Deep G-CRF~\cite{chandra2016fast} & - & 80.20  \\
    HikSeg\_COCO~\cite{sun2016mixed} & - & 81.40 \\
    SegModel~\cite{shen2017semantic} & - & 81.80 \\
    TuSimple~\cite{wang2018understanding} & - & 83.10 \\
    Large Kernel Matters~\cite{peng2017large} & 3.7G & 83.60 \\
    ResNet-38\_MS\_COCO~\cite{wu2019wider} & 12.11G & 84.90 \\
    PSPNet~\cite{zhao2017pyramid} & 16.55G & 85.40 \\
    DeepLabv3~\cite{chen2017rethinking} & 49.68G & 85.70 \\
    \specialrule{0.1pt}{0pt}{0pt}
    (Ours) - ResNet-38\_MS\_COCO & 12.25G & 85.70 \\
    (Ours) - PSPNet & 16.67 G & 86.10 \\
    \textbf{(Ours) - DeepLabv3} & 49.89G & \textbf{87.00} \\

    \specialrule{1.1pt}{0pt}{0pt}
    
  \end{tabular}
  \vspace{-0.2cm}
  \caption{Model comparison with SOTA in PASCAL-VOC-2012.}
  \label{tab:pascal}
  \vspace{-0.5cm}
\end{table}

\begin{figure}
  \centering
  \begin{subfigure}{0.232\linewidth}
    \includegraphics[width=\linewidth]{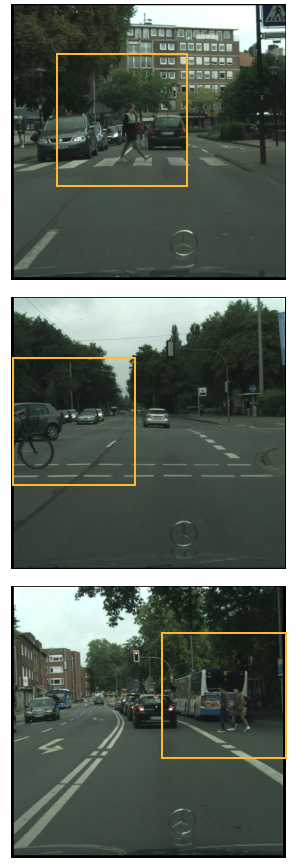}
    \caption{Image}\label{subfig:orig_city}
  \end{subfigure}
  \hfill
  \begin{subfigure}{0.232\linewidth}
    \includegraphics[width=\linewidth]{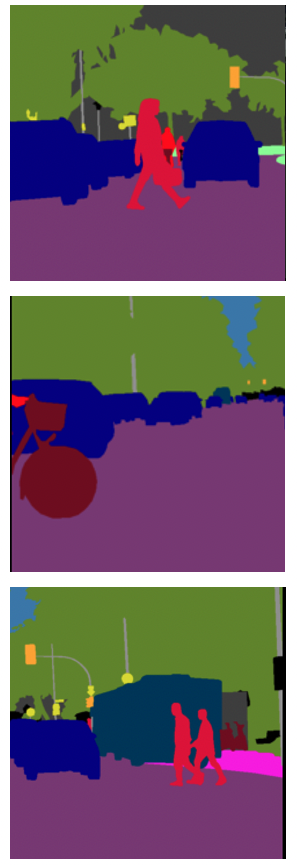}
    \caption{Ground Truth}\label{subfig:gt_city}
  \end{subfigure}
    \hfill
  \begin{subfigure}{0.231\linewidth}
    \includegraphics[width=\linewidth]{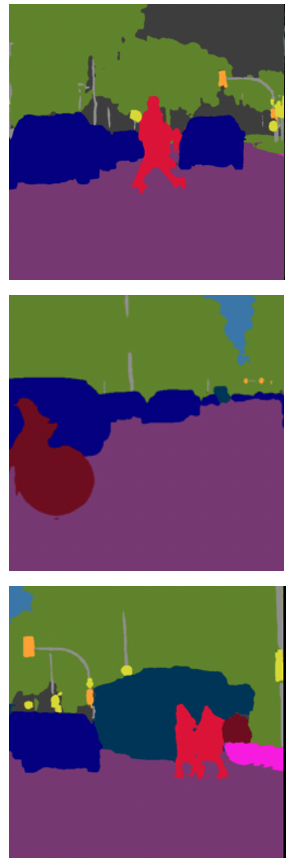}
    \caption{DeepLab-v3}\label{subfig:deep_city}
  \end{subfigure}
    \begin{subfigure}{0.23\linewidth}
    \includegraphics[width=\linewidth]{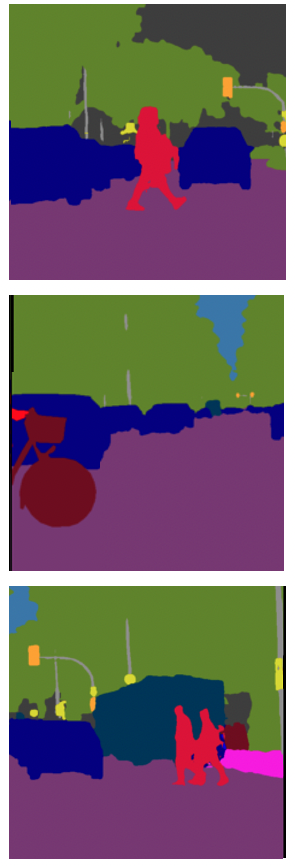}
    \caption{Ours}\label{subfig:deep_gsf_city}
  \end{subfigure}
     \hfill
    \vspace{-0.3cm}
  \caption{Zoom in of the qualitative results for Cityscapes. The fourth colum depicts our results with DAS and DeepLab-v3 integrated with the temporal shift mechanism.}
  \label{fig:city_results}
      \vspace{-0.3cm}
\end{figure}

\paragraph{Image Classification}
We use ImageNet, Cifar10, Cifar100 and TinyImageNET for the image classification experiments.~\cref{tab:ImageNet} shows the comparison of our approach with other state-of-the art models in ImageNet.
The provided results, in the lower part of the table,  are obtained running the searching procedure to find the optimal transformation for each independent backbone, \ie BNInception, Inception-v3, ResNet-50-101-152. The best found transformations are given in the Supplementary materials. We employed for this set of experiments \emph{video}-backbones trained from scratch, where we incorporate the best temporal fusion method we found with our ablation studies. 
We observe an improvement with our approach for every employed 2D backbone using roughly the same parameters, having the strongest boost of 3.15 \% in terms of accuracy-wise metric for ResNet-50 architectures. Compared to ViT-L-32 that achieves 79.66 \%, we use 1/5 of the parameters and FLOPs. In general, we observe stronger benefits for ResNet-like architectures than Inception-like ones. Moreover, as Fig.~\ref{fig:depth_ImageNet} highlights, the accuracy of such architectures, with and without our method, when reducing the depth experiences a less steep performance drop. This is reasonable and desired, given the proof of concept provided in Fig.~\ref{fig:RF}. With equal depths, we improve by a fair margin also over FFC~\cite{chi2020fast}, a popular method that achieves the expansion of the RF by employing fast fourier convolutions. 
Similar behaviours are experienced when we compare DAS+temporal expansion with other methods expanding the RF for Cifar10, Cifar100, Tiny and ImageNet datasets, in Tab~\ref{tab:c10_c100_tiny}. Of particular notice should be the drop in performance when a bigger kernel is used, probably attributable to a stronger overfitting, and the comparison with the popular dilated convolution (third column), well known for expanding the RF. We improve with respect such a method by a fair margin over each dataset. We attribute this result to the effective \emph{reshaping}, rather than enlargement of the RF.
\paragraph{Image Semantic Segmentation}
In~\cref{tab:pascal} and~\ref{tab:cityscapes} we show the performance of our method applied on Pascal-VOC-2012 and Cityscapes datasets for the image segmentation task. We compare our approach with popular methods employed in semantic segmentation, and observe a considerable gain especially for DAS combined with DeepLab (with a Resnet-101 backbone) when compared with the 2D counterpart. Indeed an improvement of 1.3 \% in the mIOU is experienced. The best set of transformations was searched for every different backbone and dataset. We indeed empirically show in Sec.~\ref{sec:ablation} how applying not optimized transformations impacts the performance.
For Cityscapes, the best results where achieved for ResNeSt backbone, with a 85.1 \% mIOU, confirming also a general trend we observed for other datasets. Indeed we experience a much higher improvement when dealing with ResNet architectures than Incpetion-like modules. This is probably due to the nature of the fusion mechanism we employ, that, when placed on the skip connections allows better preserving the spatial information. Fig.~\ref{fig:city_results} shows some qualitative results on CityScapes. One can observe a better reconstruction in details when comparing~\ref{subfig:deep_city} and~\ref{subfig:deep_gsf_city}. This is particularly visible for the reconstruction of people, (see first and third row) and of street lamps. 
We noticed an over segmentation for PASCAL-VOC-2012 dataset in some cases when applying our methodology. However our method still outperforms all compared state-of-the-art alternatives. 
We refer the reader to the Supplementary material for some qualitative results on PASCAL dataset and for additional on Cityscapes.

\begin{table}[t]
  \centering\small
  \begin{tabular}{ l >{\centering\arraybackslash}m{1.8cm} >{\centering\arraybackslash}m{1.1cm}>{\centering\arraybackslash}m{1.1cm}}
    \specialrule{1.1pt}{0pt}{0pt}
    Method  & Backbone & FLOPs & mIOU \\
    \specialrule{1.1pt}{0pt}{0pt}
    FCN~\cite{long2015fully} & ResNet-101 & - & 77.02 \\
    SETR~\cite{zheng2021rethinking} & ViT-Large & - & 78.10 \\
    Maskformer~\cite{cheng2021per} & ResNet-101 & 73G & 78.50 \\
    NonLocal~\cite{wang2018non} & ResNet-101 & - & 79.40 \\
    PSPNet~\cite{zhao2017pyramid} & ResNet-101 & 63G & 79.77 \\
    Mask2former~\cite{cheng2022masked} & ResNet-101 & - & 80.10 \\
    DeepLab-v3~\cite{chen2017rethinking} & ResNet-101 & 92.65G & 81.30\\
    DeepLab-v3~\cite{chen2017rethinking} & Xception-65 & - & 82.10 \\
    DeepLab-v3~\cite{chen2017rethinking} & ResNeSt 101 & - & 82.90\\
    DeepLab-v3~\cite{chen2017rethinking} & ResNeSt 200 & 285G & 83.30 \\
    \specialrule{0.1pt}{0pt}{0pt}
    (Ours) PSPNet & ResNet-50 & 32G & 81.21 \\
        (Ours) ResNeSt & Xception-65 & - & 82.40\\
    (Ours) DeepLab-v3 & ResNet-101 & 93G & 82.60\\
    \textbf{(Ours) ResNeSt} & \textbf{ResNet-200} & 286G & \textbf{85.10}\\
    \specialrule{1.1pt}{0pt}{0pt}
    
  \end{tabular}
  \vspace{-0.2cm}
  \caption{Modal comparison with SOTA in CityScapes.}
  \label{tab:cityscapes}
  \vspace{-0.5cm}
\end{table}

\begin{table}[h]
  \centering\small\setlength{\tabcolsep}{4pt}
  \begin{tabular}{ p{1cm} >{\centering\arraybackslash}m{1.3cm} >{\centering\arraybackslash}m{1cm} >{\centering\arraybackslash}m{1cm}>{\centering\arraybackslash}m{1cm}>{\centering\arraybackslash}m{1cm}}
    \specialrule{1.1pt}{0pt}{0pt}
    Dataset  & Model & Baseline & Dilated & $(5\times5)$ & Ours \\
    \specialrule{1.1pt}{0pt}{0pt}
     & R-18 & 94.12 & 94.31 & 92.16 & 95.12 \\
    Cifar10 & R-50  & 95.66 & 96.23 & 95.51 & 95.74\\\
     & WR-28  & 96.33 & - &  - & \textbf{96.40}\\
     \specialrule{0.1pt}{0pt}{0pt}
     
     & R-18  & 71.20 & 72.18 & 70.08 & 73.10 \\
    Cifar100 & R-50  & 74.82 & 76.12 & 75.60 & 76.71 \\
     & WR-28  & 81.40 & - & - & \textbf{83.06}\\
     \specialrule{0.1pt}{0pt}{0pt}
     
     & R-18 & 61.00 & 62.10 & 61.03 & 62.87\\
    Tiny& R-50 & 63.16 & 64.11 & 61.12 & 65.91 \\
     & R-101 & 64.21 & 65.30 & 63.60 & \textbf{67.51} \\
     \specialrule{0.1pt}{0pt}{0pt}
          & R-50 & 76.30 & 77.28 & 76.10 &  79.45\\
    ImageNET& R-101 & 77.40 & 78.15 & 76.11 & 80.05 \\
     & R-152 & 78.30 & 79.22 & 78.00 & \textbf{80.13} \\

    \specialrule{1.1pt}{0pt}{0pt}
    
  \end{tabular}
  \vspace{-0.2cm}
  \caption{Comparison in terms of accuracy with other methods expanding the RF. ``WR'' stands for WideResnet.}
  \label{tab:c10_c100_tiny}
\end{table}




  \vspace{-0.3cm}
\subsection{Ablation}\label{sec:ablation}
In this section, we present and discuss 
ablation studies
on Cifar10 and Pascal-VOC-2012. 
First, we study the effect of each component on the original task accuracy (\cref{tab:ablation_components}).  We find that the improved accuracy is not a simple matter of augmentation techniques, nor a mere contribution of the fusion mechanism. Indeed, both elements are needed as the simple stack of the same image (third column) does not imply an expansion of the RF, but actually determines a drop of performance probably due to much higher overfitting. Next, in~\cref{tab:ablation_fusion} we ablate on the best temporal fusion mechanism. As expected, we see that TSM and GSF have comparable performance, and number of parameters, with GSF being slightly better. Surprisingly, 3D CNNs perform poorly, maybe because of the small amount of data compared to the needs of 3D CNNs.
Next in~\cref{tab:ablation_augment_video} we study the impact of the auto-augmentation with our search space. We found that DAS performs best, given a fixed budget searching time of 24 hours decided a priori. We see comparable performance of the three methods for Cifar10 dataset, but we attribute such a behaviour to the relative simplicity of the task itself. Indeed, in Pascal-VOC we experience a much higher difference.
Finally, in~\cref{tab:ablation_DAS} we ablate the specificity of the found genotypes for a given dataset, providing quantitative results for the proof of concept provided in Fig.~\ref{fig:abs_fig}, \ie the effect of random transformation. We check the performance on PASCAL-VOC and Cifar-10 when the genotype to generate the video derives from the search phase for CityScapes (Random1) and for Cifar-100 (Random 2). Interestingly, we see a huge drop in performance when employing a ``random genotype", confirming the need of a carefully designed search procedure.

\begin{table}
  \centering\small
  \begin{tabular}{ l | >{\centering\arraybackslash}m{1cm} >{\centering\arraybackslash}m{1.2cm} >{\centering\arraybackslash}m{0.8cm} >{\centering\arraybackslash}m{0.7cm}}
    & Baseline & Augment & Replica & Ours \\
    \specialrule{0.1pt}{0pt}{0pt}
    Pascal-VOC & 85.40 & 85.51 & 83.40 & \textbf{86.10} \\
    Cifar-10   & 94.12 & 94.23 & 90.11 & \textbf{95.12} \\
    \specialrule{1.1pt}{0pt}{0pt}
  \end{tabular}
  \vspace{-0.2cm}
  \caption{Ablation on different model components. Augment indicates the usage of additional augmentation derived from our search space \emph{without} temporal fusion. Replica stands for the stack of image copies, that is, video produced with identity transforms and temporal fusion.} 
  \label{tab:ablation_components}
  \vspace{-0.2cm}
\end{table}

\begin{table}
  \centering\small
  \begin{tabular}{ l | >{\centering\arraybackslash}m{1.3cm} >{\centering\arraybackslash}m{1.3cm} >{\centering\arraybackslash}m{1.3cm}}
    & TSM  & 3D & GSF\\
    \specialrule{0.1pt}{0pt}{0pt}
               & 85.86  \%& 77.00 \% & \textbf{86.10 \%}\\
    Pascal-VOC &  \textbf{51.32 M} & 107 M & 51.43 M\\
               & \textbf{16.55 G}  & 26.78 G & 16.67 G\\
    \specialrule{0.1pt}{0pt}{0pt}
               & 94.77 \% & 92.10 \% & \textbf{95.12} \%\\
    Cifar-10   & \textbf{11.18 M} & 33.17 M & 11.20 M\\
               & \textbf{37.12 M }& 49 M & 37.24 M\\

    \specialrule{1.1pt}{0pt}{0pt}
  \end{tabular}
  \vspace{-0.2cm}
  \caption{Ablation on the temporal fusion method given PSPNet backbone for Pascal-VOC and Resnet-18 for Cifar10.}
  \label{tab:ablation_fusion}
  \vspace{-0.3cm}
\end{table}

\begin{table}
  \centering\small
  \begin{tabular}{ l | >{\centering\arraybackslash}m{0.8cm} >{\centering\arraybackslash}m{0.8cm} >{\centering\arraybackslash}m{0.8cm}}
    & AA & RA & DAS \\
    \specialrule{0.1pt}{0pt}{0pt}
    Pascal-VOC & 82.15 & 84.00 & \textbf{87.00}\\

    Cifar-10   & 94.35 & 92.98 & \textbf{95.12}\\

    \specialrule{1.1pt}{0pt}{0pt}
  \end{tabular}
  \vspace{-0.2cm}
  \caption{Ablation on the auto-augmentation technique given the same searching budget time.}
  \label{tab:ablation_augment_video}
  \vspace{-0.3cm}
\end{table}

\begin{table}
  \centering\small
  \begin{tabular}{ l | >{\centering\arraybackslash}m{1.6cm} >{\centering\arraybackslash}m{1.6cm} >{\centering\arraybackslash}m{0.8cm}}
    & Random 1 & Random 2 & DAS \\
    \specialrule{0.1pt}{0pt}{0pt}
    Pascal-VOC & 77.25 & 82.11 & \textbf{86.10} \\
    Cifar-10   & 93.14 & 94.15 & \textbf{95.12} \\
    \specialrule{1.1pt}{0pt}{0pt}
  \end{tabular}
  \vspace{-0.1cm}
  \caption{Ablation on the effectiveness of looking for the best transformation. Random 1 is the best transformation found for City-Scapes, Random 2 for Cifar100.}
  \label{tab:ablation_DAS}
  \vspace{-0.2cm}
\end{table}

%% file: sec/Conclusions.tex
\section{Conclusions}\label{sec:conclusions}
We proposed a differentiable auto-augmentation method for the tasks of image classification and semantic segmentation, \ie DAS. We defined a very flexible continuous search space, and employed a perturbation-based selection method to over-coming the limitations of previous approaches~\cite{cubuk2019autoaugment, cubuk2020randaugment}. We showed that by applying the optimal transformations found by DAS to generate variations of images, we can effectively reshape the RF. This is achieved by processing the new input as videos with a CNN integrated with a temporal shift mechanism that performs feature mixing in time. Our method proposes a new way of handling 2D data to exploit their richness, and investigates how the increase of the receptive field in the temporal dimension impacts the original spatial receptive field. We observed an improvement in terms of accuracy with respect to standard augmentation techniques, for both image classification and segmentation tasks, using different backbones on different datasets. We also successfully reshaped the receptive field, as shown in Fig.~\ref{subfig:RF_shapes}, which in terms of qualitative results turned out into more detailed segmentation masks.\vspace{-0.4cm}
\paragraph{Limitations and future work}Our method is compact, fast and accurate, but currently limited by its memory footprint. Memory requirement in training grows with number of generated frames, which restricted us to search transformations for 8-frame videos in our experiment. DAS with longer videos might provide further performance boosts.
Possible future work include experimenting with other backbone families, such as transformers.
Other venue of research could be the adaptation of DAS to video-to-video data augmentation, which would require a proper definition of the search space of transformations. Another interesting direction could be Deep-DAS, where optimal feature map transforms are searched at multiple network layers.

%% file: sec/X_suppl.tex
\clearpage
\setcounter{page}{1}
\setcounter{section}{0}
\maketitlesupplementary
\appendix
\section{Overview} 
This supplementary material is organized as follows. We first introduce the proof for~\cref{eq:alpha_I} and~\cref{eq:alpha_T}, which motivate our choice in the final transformations selection, based on a perturbation approach (Sec.~\ref{sec:proof_eq}). We then present the implementation details for our method, giving an overview of the hyper-parameters utilized (Sec.~\ref{sec:implementation}), and provide the details for the best transformations found by DAS for the Image-to-Video setup (Sec.~\ref{sec:DAS_celss}). We conduct an additional ablation study to further prove the effectiveness of our approach by testing the results under a re-shuffle operation, which motivates the need for the GSF as temporal shift mechanism (Sec.~\ref{sec:add_ablations}). For completeness, we report the result of DAS in the ``standard" setup of Image-to-Image, where auto-augmentation methods are usually employed and compare with two SOTA approaches, \ie AutoAugment (AA) and RandAugment (RA), conducting ablations studies to highlight the difference with respect to DAS (Sec.~\ref{sec:das_ItoI}). We conclude with more qualitative results for the semantic segmentation datasets (Sec.~\ref{sec:seg_results}) and with an additional ablation on Cifar100 testing performance robustness with reduced data (Sec.~\ref{sec:gen_reduced_data}).
\section{Proof of Equations ~\ref{eq:alpha_I}, \ref{eq:alpha_T}}
\label{sec:proof_eq}
Let $\theta_{I} = Softmax(\tau_I)$ and $\theta_{T} = Softmax(\tau_T)$. Then the mixed operation in Eq.~\ref{eq:m_bar} can be re-written as $\overline{m}(x) = \theta_{T}x_T + \theta_{I}x_I$. The objective can be formally formulated as:
\begin{equation}
    \min_{\theta_I, \theta_T} = \text{var}(\overline{m}(x) - m^*) \quad\quad s.t. \quad\quad \theta_I + \theta_T = 1
\end{equation}
\noindent
This constraint optimization problem can be solved with Lagrangian multiplies:
\begin{align}
    L(\theta_I, \theta_T, \lambda) &= \text{var}(\overline{m}(x) - m^*) - \lambda(\theta_I + \theta_T -1) \\
    &= \text{var}(\theta_{T}T(x) + \theta_{I}x - m^*) \\ \nonumber
    &\quad - \lambda(\theta_I + \theta_T -1)\\
    &= \text{var}(\theta_{T}T(x) + \theta_{I}x - (\theta_I + \theta_T)m^*) \\ \nonumber
    &\quad - \lambda(\theta_I + \theta_T -1)\\
    &= \text{var}[\theta_T(T(x) - m^*) + \theta_I(x - m^*)] \\ \nonumber
    &\quad - \lambda(\theta_I + \theta_T -1)\\
    &= \text{var}(\theta_T(T(x) -m^*) = \text{var}(x - m^*) \\ \nonumber 
    &\quad + 2\text{cov}[\theta_T(T(x)-m^*), \theta_I(x-m^*)] \\ \nonumber
    &\quad -\lambda(\theta_I + \theta_T -1)
\end{align}\vspace{-0.9cm}
\begin{align}
    &= \theta_T^2 \text{var}(T(x) - m^*) + \theta_I^2\text{var}(x-m^*) \\ \nonumber
    &\quad + 2\theta_T\theta_I\text{cov}[T(x)-m^*, x-m^*] \\ \nonumber 
    &\quad -\lambda(\theta_I + \theta_T -1)
\end{align}
Setting partial derivatives to 0
\begin{align}
    \frac{\partial L}{\partial \lambda} &= \theta_I + \theta_T -1 = 0 \\
    \frac{\partial L}{\partial \theta_T} &= 2\theta_T\text{var}(T(x) -m^*) \\ \nonumber 
    &\quad + 2\theta_I\text{cov}[T(x)-m^*, x-m^*] - \lambda = 0 \\
    \frac{\partial L}{\partial \theta_I} &= 2\theta_I\text{var}(x -m^*) \\ \nonumber
    &\quad + 2\theta_T\text{cov}[T(x)-m^*, x-m^*] - \lambda = 0
\end{align}
we obtain equations whose solution are
\begin{align}
    \theta_T\text{var}(T(x)-m^*) + \theta_I\text{cov}[T(x) -m^*, x-m^*] \\ \nonumber 
    = \theta_I\text{var}(x-m^*) + \theta_T\text{cov}[T(x)-m^*, x-m^*] 
\end{align}
Substituting $\theta_I$ with $(1-\theta_T)$ we get:
\begin{align}
    \theta_T^* = \frac{\text{var}(x-m^*) - \text{cov}[T(x)-m^*, x-m^*]}
    {z}
\end{align}
where $z = \text{var}(T(x)-m^*) + \text{var}(x-m^*) -2\text{cov}[T(x)-m^*, x-m^*]$.
Similarly we obtain
\begin{align}
    \theta_I^* = \frac{\text{var}(T(x)-m^*) - \text{cov}[(T(x)-m^*, x-m^*]}
    {z}
\end{align}
Given that $\theta_i = \frac{e^{\tau_i}}{e^{\tau_i} + e^{\tau_j}}$, with $i = I, T$, we obtain:
\begin{align}
    \tau_T^* &= \text{log}[\text{var}(x-m^*) - \text{cov}(T(x)-m^*, x-m^*)] + C\\
    \tau_I^* &= \text{log}[\text{var}(T(x)-m^*) - \text{cov}(T(x)-m^*, x-m^*)] + C    
\end{align}
where the only difference between $\tau_I$ and $\tau_T$ is the first term inside the logarithm. Therefore, if we choose the operation associated to the largest $\tau$, assuming it is related to the strength of the transformation, we will always end up choosing identity operations. This proof applies also for search spaces with more than two operations, as the transformation $T$ previously defined as a translation can be seen as the composition of multiple transformations.

\section{Implementation details}\label{sec:implementation}
Our hyper-parameters are summarized in Tab~\ref{tab:hyper}. We kept the same hyper-parameters during the search phase and the training from scratch, with the only difference in the additional optimizer needed for the Architect neural network. Such a network, responsible for the topology optimization, was trained with Adam optimizer, with $3e-4$ as learning rate and $1e-3$ as weight decay-rate. For all image datasets we applied standard augmentation techniques, such as random horizontal flip, random crop,and  cutout, on the inputs to the DAS cell. Every image, after being augmented, undergoes the temporal expansion, achieved through an image replication. Transformations are then applied inside the DAS cell to each frame so that smoothness and continuity are kept during the video generation. The image replication module acts as a ``stem" module to allow multiple cells with multiple nodes. Inference cost is not affected, as DAS is involved only during training.

\begin{table*}[h!]
  \centering
  \small
  \renewcommand{\arraystretch}{1.2} 
  \setlength{\tabcolsep}{3pt}
  \begin{tabular}{ p{4cm} >{\centering\arraybackslash}m{2cm} >{\centering\arraybackslash}m{2cm} >{\centering\arraybackslash}m{2cm}>{\centering\arraybackslash}m{2cm}>{\centering\arraybackslash}m{2cm}>{\centering\arraybackslash}m{2cm}}
    \specialrule{1.1pt}{0pt}{0pt}
      & Cifar10 & Cifar100 & Tiny & ImageNet & Pascal-VOC & CityScapes \\
    \specialrule{1.1pt}{0pt}{0pt}
    \emph{Optimization} & & & & & & \\
    Image size & (32,32) & (32,32) & (64,64) & (224,224)& (380, 380) & (1024,1024) \\
    Optimizer & SGD & SGD & SGD & SGD & SGD & SGD \\
    Batch size & 96 & 96 & 64 & 32 & 32 & 16 \\
    Learning rate scheduler & step decay & step decay & step decay & step decay & poly & poly  \\
    Base Learning rate& 0.1 & 0.1 & 0.1 & 0.1 & 0.03 & 0.03 \\
    Weight decay& 1e-4 & 1e-4 & 5e-4 & 1e-4 & 1e-4 & 1e-4 \\
    Epochs& 90 & 90 & 90 & 100 & 80 & 130\\
    \specialrule{0.1pt}{0pt}{0pt}
    Number of segments& 8 & 8 & 8 & 8 & 8 & 8\\
    \specialrule{1.1pt}{0pt}{0pt}
  \end{tabular}
  \caption{Hyperparameters employed for our experiments.}
  \label{tab:hyper}
\end{table*}

\section{Example of cells found by DAS}\label{sec:DAS_celss}

\begin{figure*}
  \centering
      \begin{subfigure}{0.48\linewidth}
  \includegraphics[width=\linewidth]{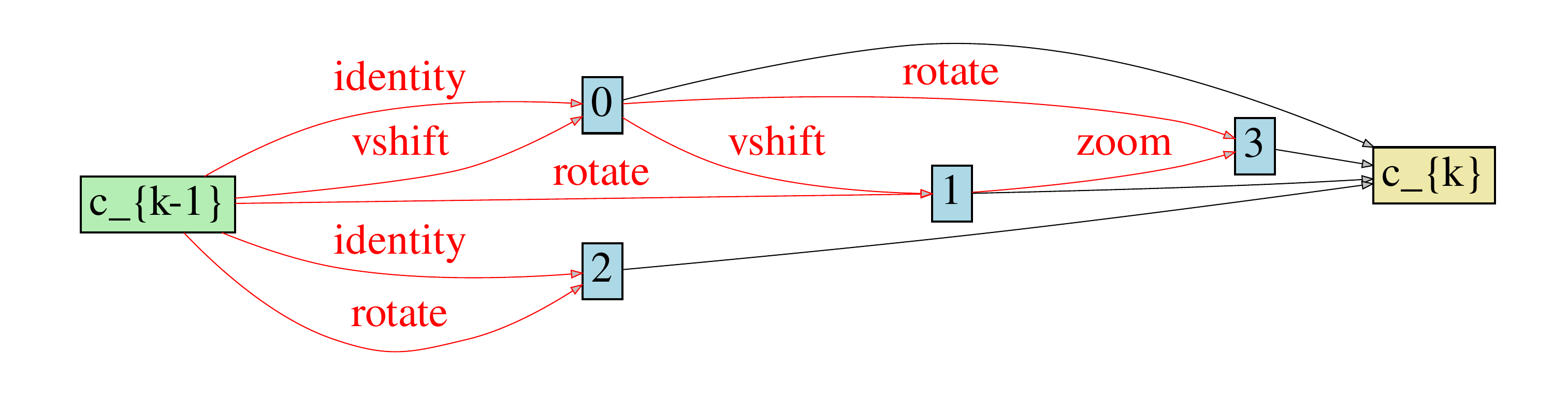}
    \caption{Cifar100.}\label{fig:c100_das}
  \end{subfigure}
  \hfill
  \begin{subfigure}{0.48\linewidth}
  \includegraphics[width=\linewidth]{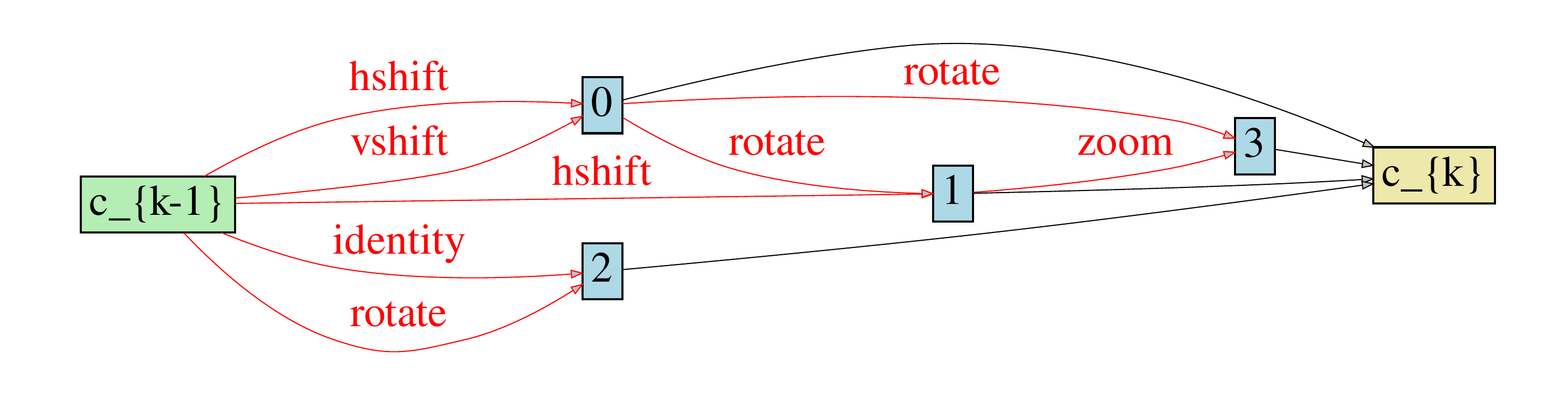}
   \caption{ImageNet.}
   \label{fig:imagenet_das}
  \end{subfigure}
  \\
  \begin{subfigure}{0.48\linewidth}
  \includegraphics[width=\linewidth]{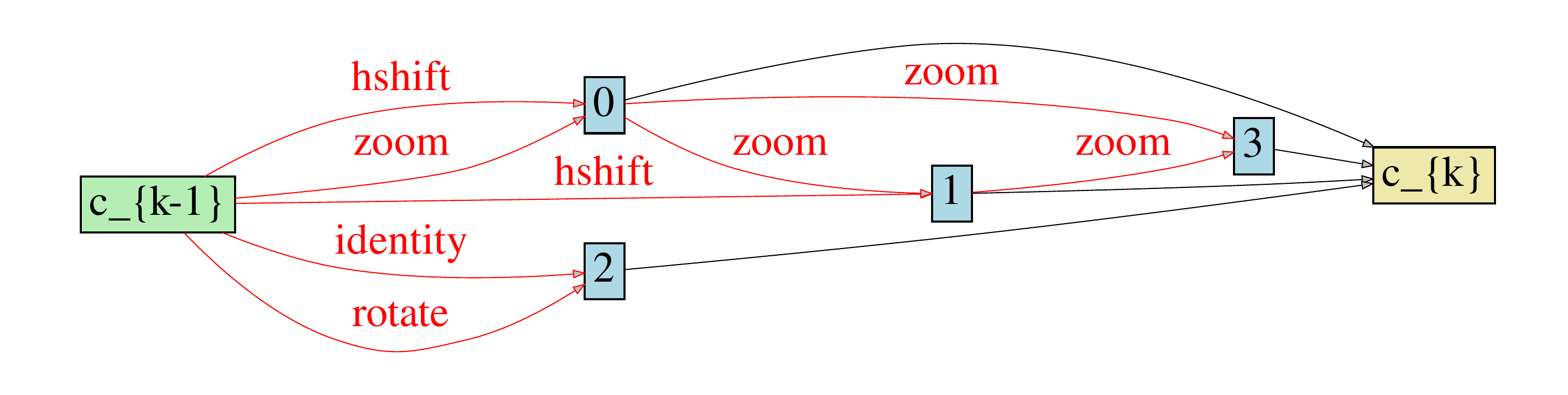}
   \caption{Pascal.}
   \label{fig:pascal_das}
  \end{subfigure}
\hfill
  \begin{subfigure}{0.48\linewidth}
  \includegraphics[width=\linewidth]{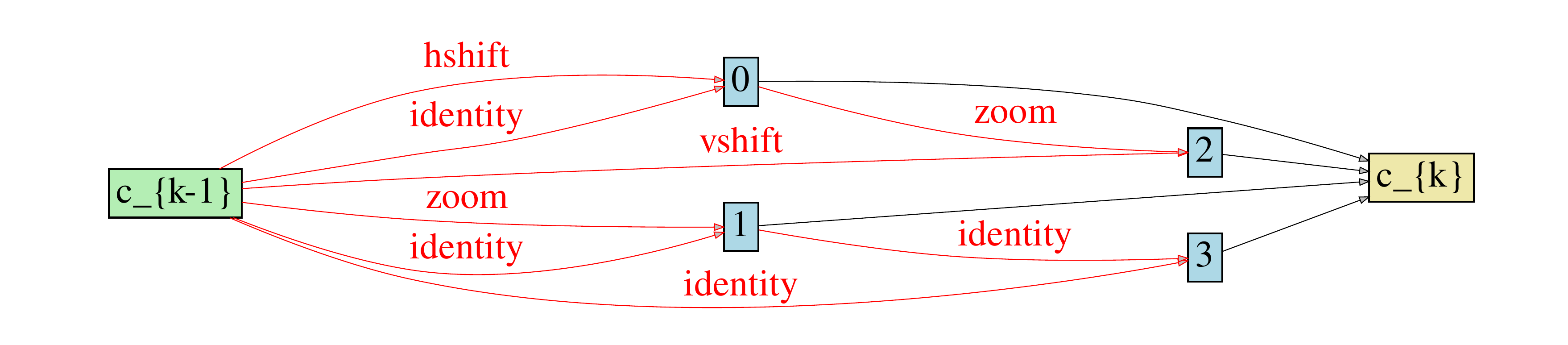}
   \caption{Cityscapes.}
   \label{fig:city_das}
  \end{subfigure}
     \hfill
     \vspace{-0.3cm}
  \caption{Best transformations found by DAS.}
  \label{fig:das_results}
      \vspace{-0.3cm}
\end{figure*}

We provide the graph visualization for the cells found by DAS for Cifar100 (Fig.~\ref{fig:c100_das}), ImageNet (Fig.~\ref{fig:imagenet_das}), Pascal-VOC (Fig.~\ref{fig:pascal_das}) and Cityscapes (Fig.~\ref{fig:city_das}). We do not report the results for Cifar10 and Tiny ImageNet as the found cell is the same as Cifar100 and ImageNet, respectively. This justifies the results previously introduced in Tab.~\ref{tab:ablation_DAS} for Cifar-10.

\section{Additional ablations on DAS for Image-to-Video}\label{sec:add_ablations}
\begin{figure}[t]
  \centering
    \includegraphics[width=8cm, height=4cm]{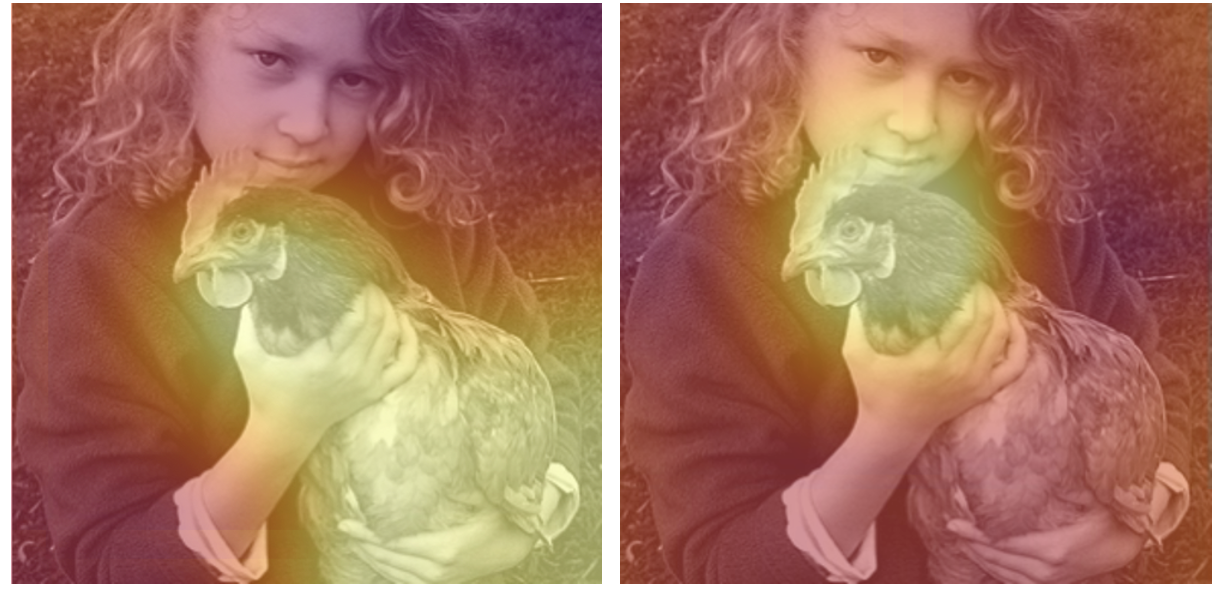}
    \caption{Receptive Field shape difference between our method (left) and standard 2D CNNs (right).}
    \label{fig:RF_more_new}
\end{figure}

Tab.~\ref{tab:extra_ablations} compares the results we previously showed in Tab.~\ref{tab:ablation_components} in the main paper, with an additional experiment to prove the need for the GSF component. To this aim, the frames of the video input (obtained with the best transformations found by DAS) are randomly shuffled with the goal of loosing the temporal continuity. This experiment aims at showing that both components, DAS and GSF, are needed, but does not imply a limitation of DAS in the search space definition. As the optimization of the DAS cell to find the optimal transformations occurs during the training of the network, even given a huge search space with non continuous transformations, DAS will optimize to find the best transformations that lead to the highest validation accuracy for \emph{that} architecture. As a result, as we show with further experiments in Sec.~\ref{sec:das_ItoI} the approach stays robust even under noisy transformations. The experiments are run with a PSPNet with ResNet-50 backbone for Pascal-VOC dataset and with ResNet-18 for Cifar10 dataset. For each dataset, we show the accuracy (first row) the \# of parameters (second row) and the number of flops (third row) with an input size $32\times32$ and $400\times400$ for Cifar10 and Pascal-VOC, respectively. Finally,~\cref{fig:RF_more_new} gives an example of our RF (left) and standard 2d CNN (right) for an ImageNet sample.

\begin{table}
  \centering
  \footnotesize
  \begin{tabular}{ p{1cm} | >{\centering\arraybackslash}m{0.9cm} >{\centering\arraybackslash}m{1.6cm}| >{\centering\arraybackslash}m{1.6cm} >{\centering\arraybackslash}m{0.6cm}}
    & Baseline & DAS Aug (\emph{S})&Re-shuffle & Ours \\
    \specialrule{0.1pt}{0pt}{0pt}
    Pascal & 85.40 & 85.51 &  85.44 & 86.10 \\
        \specialrule{0.1pt}{0pt}{0pt}
           &  \multicolumn{2}{c|}{51.32 M}  &  \multicolumn{2}{c}{51.43 M}      \\
            &  \multicolumn{2}{c|}{16.55 Gflops}  &  \multicolumn{2}{c}{16.67 Gflops}      \\
                \specialrule{0.1pt}{0pt}{0pt}
    Cifar10   & 94.12 & 94.23 & 94.15 & 95.12\\
        \specialrule{0.1pt}{0pt}{0pt}
               &  \multicolumn{2}{c|}{11.18 M}  &  \multicolumn{2}{c}{11.20 M}      \\
            &  \multicolumn{2}{c|}{37.12 Mflops}  &  \multicolumn{2}{c}{37.12 Mflops}      \\
    \specialrule{1.1pt}{0pt}{0pt}
  \end{tabular}
  \caption{Additional ablation experiments. Baseline was obtained with the 2D backbone with standard augmentation techniques. ``DAS Aug (\emph{S})" stands for the inclusion of additional DAS augmentations in Space \emph{S}, meaning that the data is processed by a 2D backbone. Re-shuffle processes the input in the same way as DAS Aug (\emph{S}) but stacks the transformations in the temporal dimension to create a video, and subsequently re-shuffles the frames of the video. ``Ours" processes the input obtained with DAS with temporal continuity preserved. The backbone for the last two experiments is 2D+temporal shift.}
  \label{tab:extra_ablations}
  \vspace{-0.5cm}
\end{table}

The little difference in the ``re-shuffle" experiment performed for Pascal-VOC and Cifar-10 datasets with respect to the baseline and DAS Aug \emph{S} is probably due to perturbations. The temporal shift mechanism, \ie GSF,  is designed to learn to shift features among adjacent frames. However, if those features are not consistent across the time dimension, GSF correctly learns not to route gated features.  As a result, the experiment reconducts to processing data augemnted as in DAS Aug \emph{S} with a 2D backbone integrated with a temporal shift mechanism that learns not to shift.

\section{Experiments on DAS for Image-to-Image}\label{sec:das_ItoI}
\subsection{Comparison with SOTAs} 
Tab.~\ref{tab:das_itoi} compares our Differentiable Augmentation Search with other SOTA auto-augmentation techniques, \ie AA [1] and RA [2] for the task of image-to-image. This means that no temporal expansion is performed, and a comparable search-space usually deployed for finding standard data-augmentation is defined. Similar to AA and RA, we define in our search space the following set of transformations: Shear X/Y, Translate X/Y, Rotate, AutoContrast, Invert, Equalize, Solarize, Posterize, Color, Brightness, Sharpness, Cutout, and Identity that corresponds to applying no transformation. We run experiments on Cifar-10, Cifar-100, SVHN, and ImageNet, for this set of experiments, we did not fix a budget time for the required search time. Following RA setup, for comparison purposes, we employed a Wide-ResNet-28-2 for the first three datasets, and a ResNet-50 model for ImageNET. 
\begin{table}
  \centering
  \footnotesize
  \begin{tabular}{ >{\centering\arraybackslash}m{1.25cm} | >{\raggedright\arraybackslash}m{0.6cm}| >{\centering\arraybackslash}m{1.0cm} | >{\centering\arraybackslash}m{1.2cm}| >{\centering\arraybackslash}m{0.8cm} | >{\centering\arraybackslash}m{1.1cm}}
    & \centering search & Cifar-10 & Cifar-100 & SVHN & ImageNet \\
    & \centering space & WRN & WRN & WRN & ResNet \\
    \specialrule{0.1pt}{0pt}{0pt}
Baseline & \centering 0 & 94.90 & 75.40 & 96.70 & 76.30 \\
  \specialrule{0.1pt}{0pt}{0pt}
AA & $10^{32}$ & 95.90 & 78.50  & 98.00& 77.60 \\
RA & $10^2$& 95.80 & 78.30 & \textbf{98.30} & 77.60 \\
DAS & $10^{13}$ & \textbf{96.10} & \textbf{78.90} &\textbf{98.30} & \textbf{77.90 }\\
\end{tabular}
  \caption{Comparison among different auto-augmentation methods. WRN stands for Wide-ResNet-28-2, while ResNet is the ResNet-50 model. Best results are bolded.}
  \label{tab:das_itoi}
\end{table}

DAS out-performs previous auto-augmentation methods in all datasets but SVHN, where it equals RA performance.  
\subsection{Advantages of DAS}
\begin{figure}[h!]
  \centering
   \includegraphics[width=0.95\linewidth]{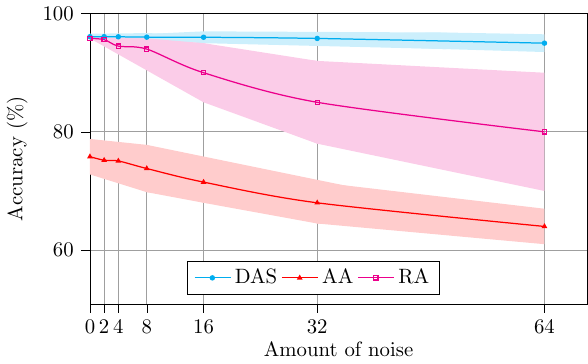}
  \vspace{-0.3cm}
   \caption{Results on Cifar-10 for each auto-augmentation technique. Experiments are run 5 times, with the shaded area representing the variance.}
   \label{fig:ablation_dasitoi}
\end{figure}

We ablate now on the importance of introducing our differentiable algorithm highlighting the two main drawbacks of the cited competitors. On the one hand, AA is extremely competitive in terms of obtained accuracy, surpassing RA in Cifar-10, Cifar-100, and having equal performance on Imagenet. However, AA is extremely slow, requiring 15000 GPU hours to look for the optimal policy on a \emph{reduced} ImageNet. On the other hand RA is extremely efficient, as it reduces the search space to $10^2$ different choices, but we argue it is not robust when introducing not relevant transformations. The authors of [2] indeed show that when introducing color transformations in the Cifar-10 experiments, they experience a degradation of validation accuracy on average. This implies that one needs to carefully design the search space, and cannot include transformations that potentially may harm the performance on the dataset. A justification for such a behaviour is due to their search space definition, where a transformation is selected with uniform probability $1/K$. This implies that as the number of K transformations in the search space increases, the probability is reduced, and the time required to find the best transformations increases. On the other hand, under a fixed searching-time budget, this results in a higher variability when the procedure is run multiple times. Evidence supporting this is displayed in Fig.~\ref{fig:ablation_dasitoi}, where we fix for Cifar-10 a searching budget time of 24 hours, and exacerbate this behaviour by progressively adding a noise transformation.\\
  \begin{figure}[t]
  \centering
    \includegraphics[width=8cm, height=4.5cm]{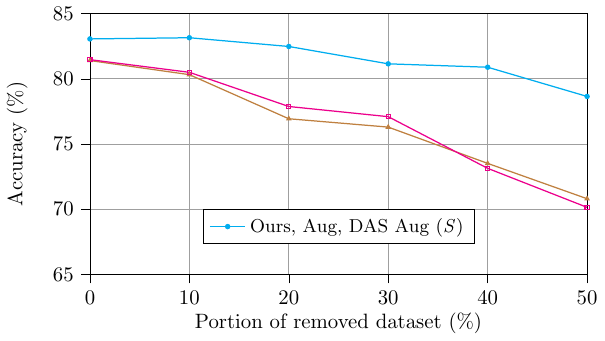}
    \caption{Top-1 test accuracy on Cifar-100 dataset given different portion of removed training dataset.}
    \label{fig:new_reduced}
\end{figure}
\section{Segmentation results}\label{sec:seg_results}
\begin{figure*}[h!]
  \centering
      \begin{subfigure}{0.7\linewidth}
  \includegraphics[width=\linewidth]{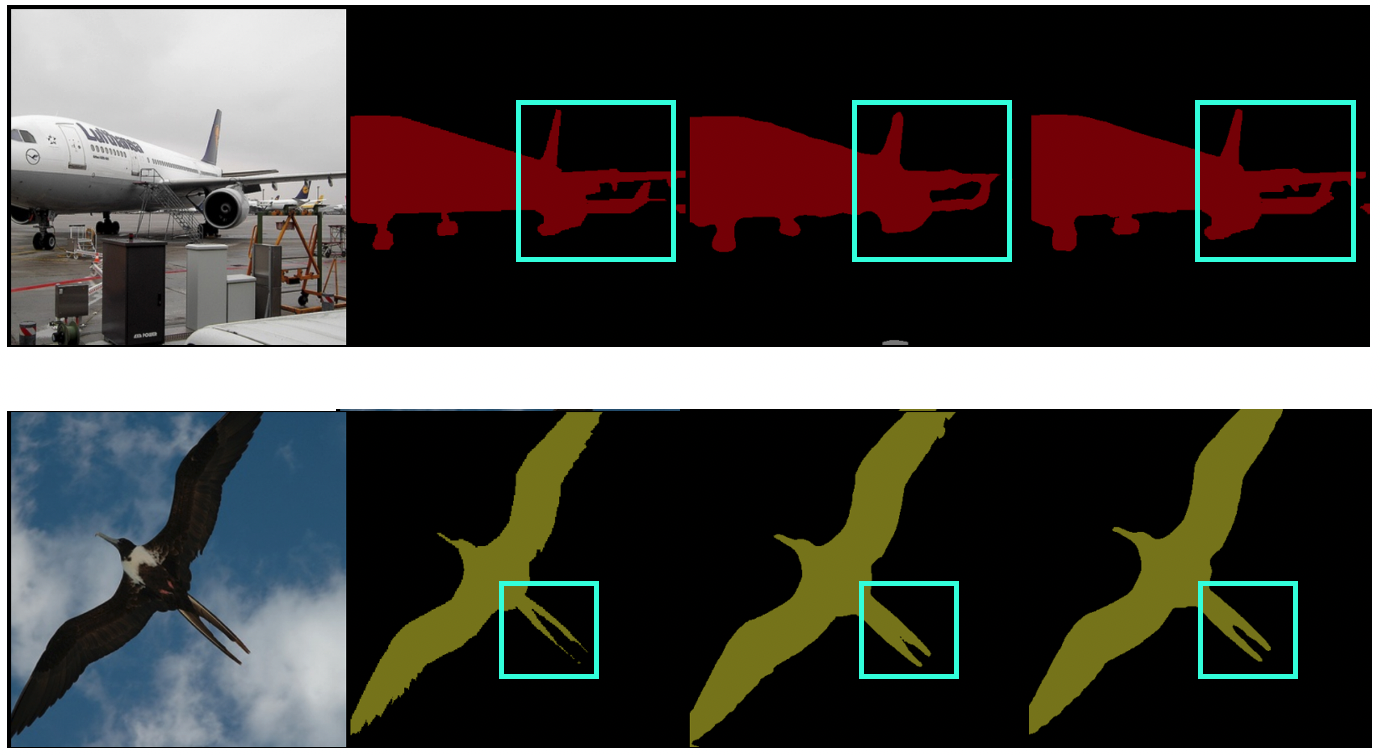}
    \label{fig:voc1}
  \end{subfigure}
  \\
  \begin{subfigure}{0.7\linewidth}
  \includegraphics[width=\linewidth]{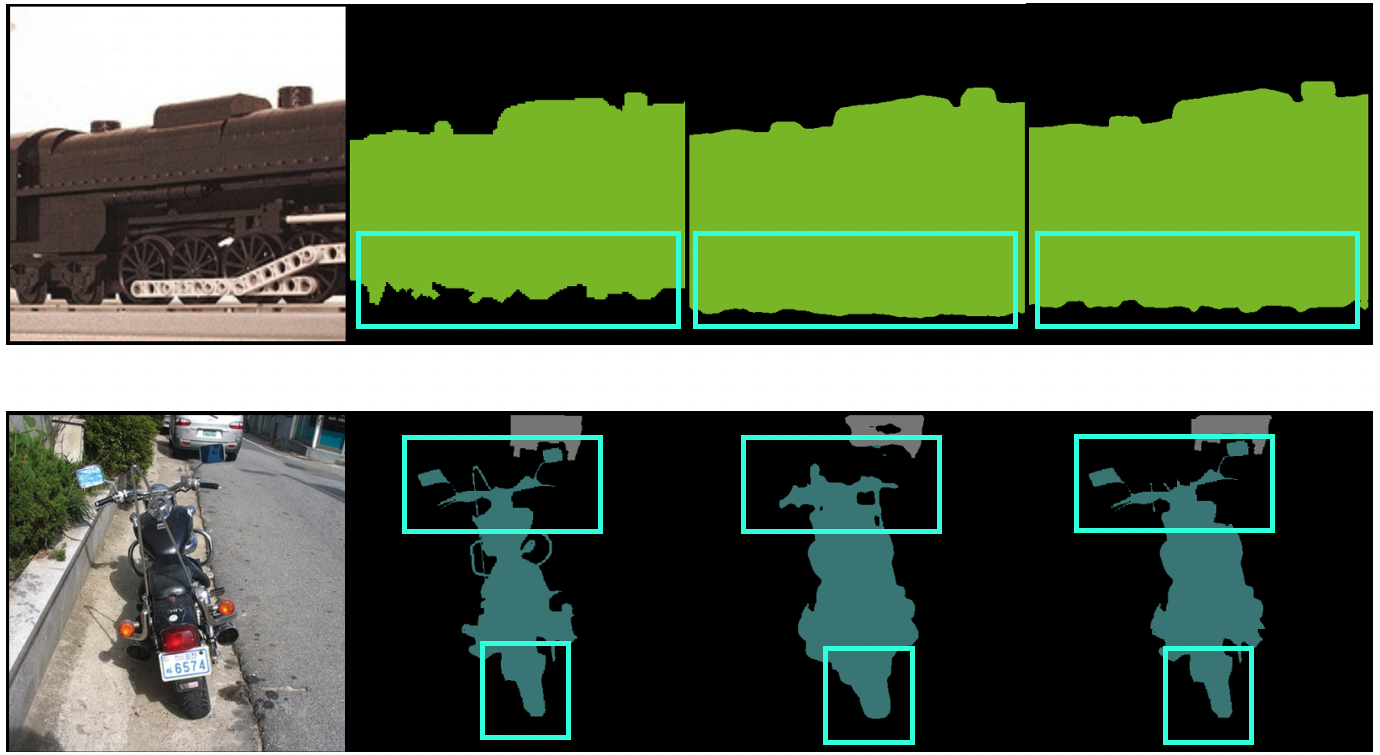}
   \label{fig:voc2}
  \end{subfigure}
  \\
    \begin{subfigure}{0.7\linewidth}
  \includegraphics[width=\linewidth]{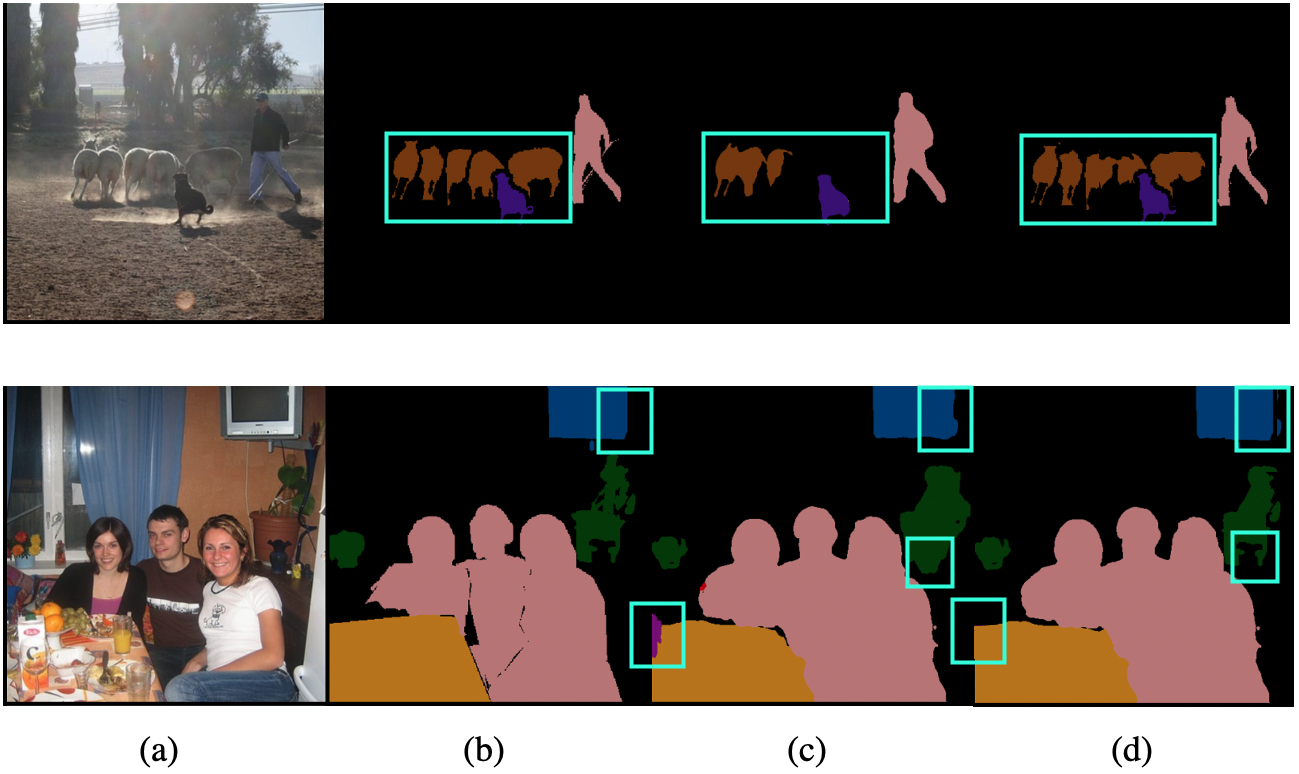}
   \label{fig:voc3}
  \end{subfigure}
 
  \caption{VOC qualitative results. Original image (a), Ground Truth (b), DeepLabv3 (c) and Ours (d) images are displayed.}
  \label{fig:voc_results}
\end{figure*}

\begin{figure*}
  \centering
      \begin{subfigure}{0.69\linewidth}
  \includegraphics[width=\linewidth]{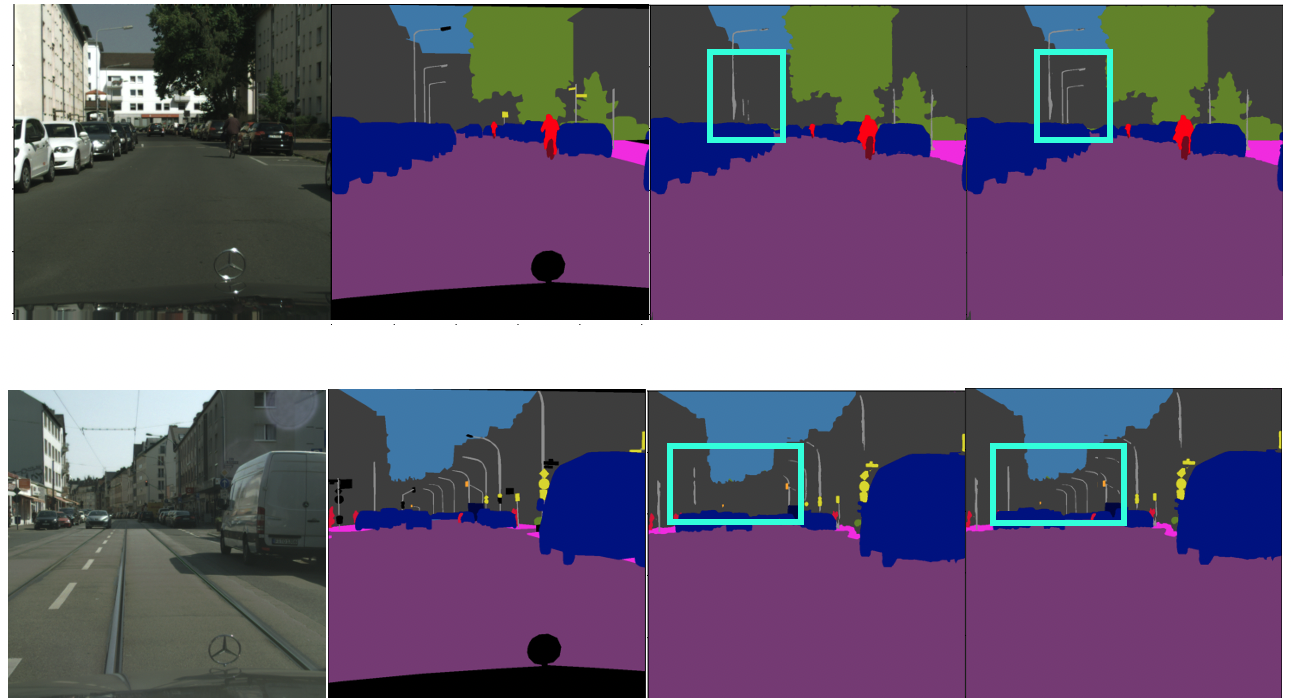}
   \label{fig:city1}
  \end{subfigure}
  \\
  \begin{subfigure}{0.7\linewidth}
  \includegraphics[width=\linewidth]{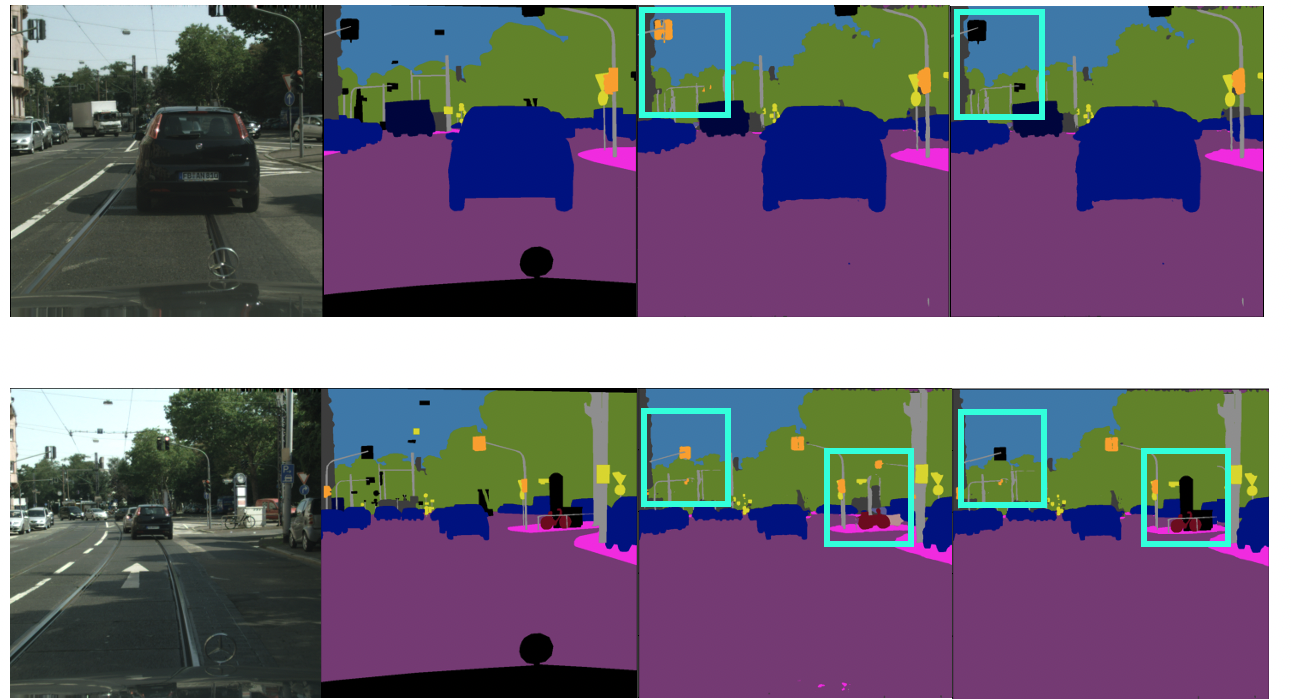}
   \label{fig:city2}
  \end{subfigure}
 \\
   \begin{subfigure}{0.7\linewidth}
  \includegraphics[width=\linewidth]{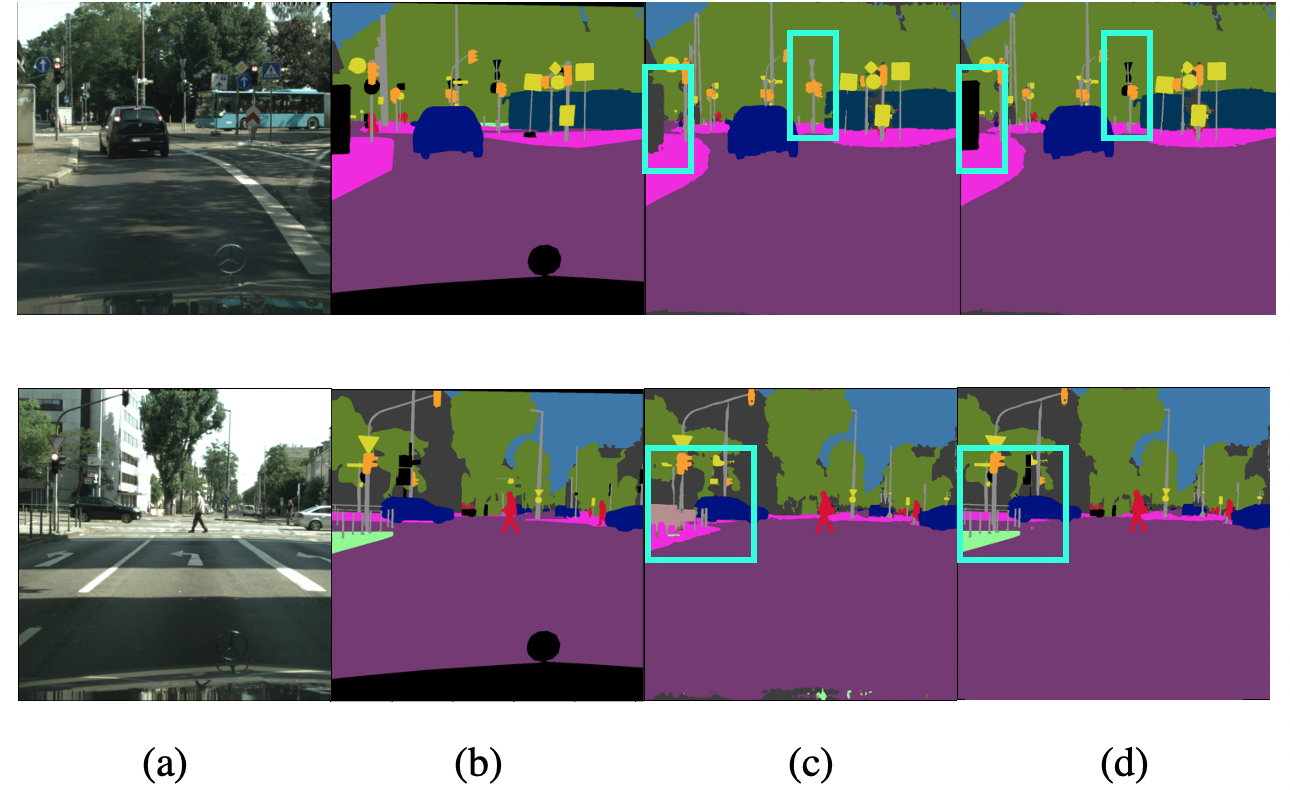}
   \label{fig:city3}
  \end{subfigure}
  \caption{City qualitative results. Original image (a), Ground Truth (b), DeepLabv3 (c) and Ours (d) images are displayed.}
  \label{fig:city_resultss}
\end{figure*}

We provide more segmentation results on Pascal (Fig.~\ref{fig:voc_results}) and CityScapes (Fig.~\ref{fig:city_resultss}) datasets. In our the figures we provide the original image (first column), the ground truth (second column), results from DeepLabv3 (column 3) and results with our methods (column 4). We highlight with a square the details where attention should be put to appreciate the difference in the results. We observe in our method, as general behaviour, a stronger capability in reconstructing details, \eg the back part of the airplane, the details in the motorcycle, plants in Pascal-VOC, street lamps in Cityscapes. We also see that, with respect to the baseline, fewer classes are misclassified, as it can be seen for the portion of the table in the sixth row of Pascal-VOC results, in traffic lights in the third row of Cityscapes results, and in the sidewalk of the sixth row of Cityscapes.

\section{Generalizability with reduced training data}\label{sec:gen_reduced_data}
To strengthen our point we run a further ablation on Cifar100, shown in~\cref{fig:new_reduced}. When reducing the size of the dataset, we barely experience a performance degradation (compared to standard augmentations (Aug) and to DAS augmnentations not concatenated in time (DAS Aug \emph{S})), finding a very useful application in scenarios where few data are available. Compared to finding new data, the cost of representing an image as a video is largely reduced.

%% file: main.bbl
\begin{thebibliography}{64}
\providecommand{\natexlab}[1]{#1}
\providecommand{\url}[1]{\texttt{#1}}
\expandafter\ifx\csname urlstyle\endcsname\relax
  \providecommand{\doi}[1]{doi: #1}\else
  \providecommand{\doi}{doi: \begingroup \urlstyle{rm}\Url}\fi

\bibitem[Araujo et~al.(2019)Araujo, Norris, and Sim]{araujo2019computing}
André Araujo, Wade Norris, and Jack Sim.
\newblock Computing receptive fields of convolutional neural networks.
\newblock \emph{Distill}, 2019.
\newblock https://distill.pub/2019/computing-receptive-fields.

\bibitem[Chandra and Kokkinos(2016)]{chandra2016fast}
Siddhartha Chandra and Iasonas Kokkinos.
\newblock Fast, exact and multi-scale inference for semantic image segmentation with deep gaussian crfs.
\newblock In \emph{Computer Vision--ECCV 2016: 14th European Conference, Amsterdam, The Netherlands, October 11--14, 2016, Proceedings, Part VII 14}, pages 402--418. Springer, 2016.

\bibitem[Chen et~al.(2017)Chen, Papandreou, Schroff, and Adam]{chen2017rethinking}
Liang-Chieh Chen, George Papandreou, Florian Schroff, and Hartwig Adam.
\newblock Rethinking atrous convolution for semantic image segmentation.
\newblock \emph{arXiv preprint arXiv:1706.05587}, 2017.

\bibitem[Chen et~al.(2018{\natexlab{a}})Chen, Papandreou, Kokkinos, Murphy, and Yuille]{Chen_2018}
Liang-Chieh Chen, George Papandreou, Iasonas Kokkinos, Kevin Murphy, and Alan~L. Yuille.
\newblock Deeplab: Semantic image segmentation with deep convolutional nets, atrous convolution, and fully connected crfs.
\newblock \emph{IEEE Transactions on Pattern Analysis and Machine Intelligence}, 40\penalty0 (4):\penalty0 834--848, 2018{\natexlab{a}}.

\bibitem[Chen et~al.(2018{\natexlab{b}})Chen, Zhu, Papandreou, Schroff, and Adam]{chen2018encoder}
Liang-Chieh Chen, Yukun Zhu, George Papandreou, Florian Schroff, and Hartwig Adam.
\newblock Encoder-decoder with atrous separable convolution for semantic image segmentation.
\newblock In \emph{Proceedings of the European conference on computer vision (ECCV)}, pages 801--818, 2018{\natexlab{b}}.

\bibitem[Chen et~al.(2019)Chen, Fan, Xu, Yan, Kalantidis, Rohrbach, Yan, and Feng]{chen2019drop}
Yunpeng Chen, Haoqi Fan, Bing Xu, Zhicheng Yan, Yannis Kalantidis, Marcus Rohrbach, Shuicheng Yan, and Jiashi Feng.
\newblock Drop an octave: Reducing spatial redundancy in convolutional neural networks with octave convolution.
\newblock In \emph{Proceedings of the IEEE/CVF international conference on computer vision}, pages 3435--3444, 2019.

\bibitem[Cheng et~al.(2021)Cheng, Schwing, and Kirillov]{cheng2021per}
Bowen Cheng, Alex Schwing, and Alexander Kirillov.
\newblock Per-pixel classification is not all you need for semantic segmentation.
\newblock \emph{Advances in Neural Information Processing Systems}, 34:\penalty0 17864--17875, 2021.

\bibitem[Cheng et~al.(2022)Cheng, Misra, Schwing, Kirillov, and Girdhar]{cheng2022masked}
Bowen Cheng, Ishan Misra, Alexander~G Schwing, Alexander Kirillov, and Rohit Girdhar.
\newblock Masked-attention mask transformer for universal image segmentation.
\newblock In \emph{Proceedings of the IEEE/CVF conference on computer vision and pattern recognition}, pages 1290--1299, 2022.

\bibitem[Chi et~al.(2020)Chi, Jiang, and Mu]{chi2020fast}
Lu Chi, Borui Jiang, and Yadong Mu.
\newblock Fast fourier convolution.
\newblock \emph{Advances in Neural Information Processing Systems}, 33:\penalty0 4479--4488, 2020.

\bibitem[Colson et~al.(2007)Colson, Marcotte, and Savard]{colson2007overview}
Beno{\^\i}t Colson, Patrice Marcotte, and Gilles Savard.
\newblock An overview of bilevel optimization.
\newblock \emph{Annals of operations research}, 153:\penalty0 235--256, 2007.

\bibitem[Cubuk et~al.(2019)Cubuk, Zoph, Mane, Vasudevan, and Le]{cubuk2019autoaugment}
Ekin~D Cubuk, Barret Zoph, Dandelion Mane, Vijay Vasudevan, and Quoc~V Le.
\newblock Autoaugment: Learning augmentation strategies from data.
\newblock In \emph{Proceedings of the IEEE/CVF conference on computer vision and pattern recognition}, pages 113--123, 2019.

\bibitem[Cubuk et~al.(2020)Cubuk, Zoph, Shlens, and Le]{cubuk2020randaugment}
Ekin~D Cubuk, Barret Zoph, Jonathon Shlens, and Quoc~V Le.
\newblock Randaugment: Practical automated data augmentation with a reduced search space.
\newblock In \emph{Proceedings of the IEEE/CVF conference on computer vision and pattern recognition workshops}, pages 702--703, 2020.

\bibitem[Dosovitskiy et~al.(2020)Dosovitskiy, Beyer, Kolesnikov, Weissenborn, Zhai, Unterthiner, Dehghani, Minderer, Heigold, Gelly, et~al.]{dosovitskiy2020image}
Alexey Dosovitskiy, Lucas Beyer, Alexander Kolesnikov, Dirk Weissenborn, Xiaohua Zhai, Thomas Unterthiner, Mostafa Dehghani, Matthias Minderer, Georg Heigold, Sylvain Gelly, et~al.
\newblock An image is worth 16x16 words: Transformers for image recognition at scale.
\newblock In \emph{International Conference on Learning Representations}, 2020.

\bibitem[Fu et~al.(2019)Fu, Liu, Tian, Li, Bao, Fang, and Lu]{fu2019dual}
Jun Fu, Jing Liu, Haijie Tian, Yong Li, Yongjun Bao, Zhiwei Fang, and Hanqing Lu.
\newblock Dual attention network for scene segmentation.
\newblock In \emph{Proceedings of the IEEE/CVF conference on computer vision and pattern recognition}, pages 3146--3154, 2019.

\bibitem[Gao et~al.(2019)Gao, Cheng, Zhao, Zhang, Yang, and Torr]{gao2019res2net}
Shang-Hua Gao, Ming-Ming Cheng, Kai Zhao, Xin-Yu Zhang, Ming-Hsuan Yang, and Philip Torr.
\newblock Res2net: A new multi-scale backbone architecture.
\newblock \emph{IEEE transactions on pattern analysis and machine intelligence}, 43\penalty0 (2):\penalty0 652--662, 2019.

\bibitem[He et~al.(2016)He, Zhang, Ren, and Sun]{he2016deep}
Kaiming He, Xiangyu Zhang, Shaoqing Ren, and Jian Sun.
\newblock Deep residual learning for image recognition.
\newblock In \emph{Proceedings of the IEEE conference on computer vision and pattern recognition}, pages 770--778, 2016.

\bibitem[Ho et~al.(2019)Ho, Liang, Chen, Stoica, and Abbeel]{ho2019population}
Daniel Ho, Eric Liang, Xi Chen, Ion Stoica, and Pieter Abbeel.
\newblock Population based augmentation: Efficient learning of augmentation policy schedules.
\newblock In \emph{International conference on machine learning}, pages 2731--2741. PMLR, 2019.

\bibitem[Howard et~al.(2017)Howard, Zhu, Chen, Kalenichenko, Wang, Weyand, Andreetto, and Adam]{howard2017mobilenets}
Andrew~G Howard, Menglong Zhu, Bo Chen, Dmitry Kalenichenko, Weijun Wang, Tobias Weyand, Marco Andreetto, and Hartwig Adam.
\newblock Mobilenets: Efficient convolutional neural networks for mobile vision applications.
\newblock \emph{arXiv preprint arXiv:1704.04861}, 2017.

\bibitem[Hu et~al.(2018)Hu, Shen, and Sun]{hu2018squeeze}
Jie Hu, Li Shen, and Gang Sun.
\newblock Squeeze-and-excitation networks.
\newblock In \emph{Proceedings of the IEEE conference on computer vision and pattern recognition}, pages 7132--7141, 2018.

\bibitem[Ioffe and Szegedy(2015)]{ioffe2015batch}
Sergey Ioffe and Christian Szegedy.
\newblock Batch normalization: Accelerating deep network training by reducing internal covariate shift.
\newblock In \emph{International conference on machine learning}, pages 448--456. pmlr, 2015.

\bibitem[Jian et~al.(2016)Jian, Kaiming, Shaoqing, and Xiangyu]{jian2016deep}
S Jian, H Kaiming, R Shaoqing, and Z Xiangyu.
\newblock Deep residual learning for image recognition.
\newblock In \emph{IEEE Conference on Computer Vision \& Pattern Recognition}, pages 770--778, 2016.

\bibitem[Krizhevsky et~al.(2012)Krizhevsky, Sutskever, and Hinton]{krizhevsky2012imagenet}
Alex Krizhevsky, Ilya Sutskever, and Geoffrey~E Hinton.
\newblock Imagenet classification with deep convolutional neural networks.
\newblock \emph{Advances in neural information processing systems}, 25, 2012.

\bibitem[LeCun et~al.(1998)LeCun, Bottou, Bengio, and Haffner]{lecun1998gradient}
Yann LeCun, L{\'e}on Bottou, Yoshua Bengio, and Patrick Haffner.
\newblock Gradient-based learning applied to document recognition.
\newblock \emph{Proceedings of the IEEE}, 86\penalty0 (11):\penalty0 2278--2324, 1998.

\bibitem[Lim et~al.(2019)Lim, Kim, Kim, Kim, and Kim]{lim2019fast}
Sungbin Lim, Ildoo Kim, Taesup Kim, Chiheon Kim, and Sungwoong Kim.
\newblock Fast autoaugment.
\newblock \emph{Advances in Neural Information Processing Systems}, 32, 2019.

\bibitem[Lin et~al.(2019{\natexlab{a}})Lin, Guo, Li, Yuan, Wu, Yan, Lin, and Ouyang]{lin2019online}
Chen Lin, Minghao Guo, Chuming Li, Xin Yuan, Wei Wu, Junjie Yan, Dahua Lin, and Wanli Ouyang.
\newblock Online hyper-parameter learning for auto-augmentation strategy.
\newblock In \emph{Proceedings of the IEEE/CVF international conference on computer vision}, pages 6579--6588, 2019{\natexlab{a}}.

\bibitem[Lin et~al.(2019{\natexlab{b}})Lin, Gan, and Han]{Lin_2019}
Ji Lin, Chuang Gan, and Song Han.
\newblock Tsm: Temporal shift module for efficient video understanding.
\newblock In \emph{Proceedings of the IEEE/CVF International Conference on Computer Vision}, pages 7083--7093, 2019{\natexlab{b}}.

\bibitem[LingChen et~al.(2020)LingChen, Khonsari, Lashkari, Nazari, Sambee, and Nascimento]{lingchen2020uniformaugment}
Tom~Ching LingChen, Ava Khonsari, Amirreza Lashkari, Mina~Rafi Nazari, Jaspreet~Singh Sambee, and Mario~A Nascimento.
\newblock Uniformaugment: A search-free probabilistic data augmentation approach.
\newblock \emph{arXiv preprint arXiv:2003.14348}, 2020.

\bibitem[Liu et~al.(2018)Liu, Simonyan, and Yang]{liu2018darts}
Hanxiao Liu, Karen Simonyan, and Yiming Yang.
\newblock Darts: Differentiable architecture search.
\newblock In \emph{International Conference on Learning Representations}, 2018.

\bibitem[Liu et~al.(2016)Liu, Rabinovich, and Berg]{Liu2015ParseNetLW}
Wei Liu, Andrew Rabinovich, and Alexander~C. Berg.
\newblock Parsenet: Looking wider to see better.
\newblock abs/1506.04579, 2016.

\bibitem[Liu et~al.(2021)Liu, Lin, Cao, Hu, Wei, Zhang, Lin, and Guo]{swin}
Ze Liu, Yutong Lin, Yue Cao, Han Hu, Yixuan Wei, Zheng Zhang, Stephen Lin, and Baining Guo.
\newblock Swin transformer: Hierarchical vision transformer using shifted windows.
\newblock \emph{2021 IEEE/CVF International Conference on Computer Vision (ICCV)}, pages 9992--10002, 2021.

\bibitem[Long et~al.(2015)Long, Shelhamer, and Darrell]{long2015fully}
Jonathan Long, Evan Shelhamer, and Trevor Darrell.
\newblock Fully convolutional networks for semantic segmentation.
\newblock In \emph{Proceedings of the IEEE conference on computer vision and pattern recognition}, pages 3431--3440, 2015.

\bibitem[Luo and Yuille(2019)]{Luo_2019}
Chenxu Luo and Alan~L Yuille.
\newblock Grouped spatial-temporal aggregation for efficient action recognition.
\newblock In \emph{Proceedings of the IEEE/CVF International Conference on Computer Vision}, pages 5512--5521, 2019.

\bibitem[Mostajabi et~al.(2015)Mostajabi, Yadollahpour, and Shakhnarovich]{Mostajabi_2014}
Mohammadreza Mostajabi, Payman Yadollahpour, and Gregory Shakhnarovich.
\newblock Feedforward semantic segmentation with zoom-out features.
\newblock In \emph{2015 IEEE Conference on Computer Vision and Pattern Recognition (CVPR)}. IEEE, 2015.

\bibitem[M{\"u}ller and Hutter(2021)]{muller2021trivialaugment}
Samuel~G M{\"u}ller and Frank Hutter.
\newblock Trivialaugment: Tuning-free yet state-of-the-art data augmentation.
\newblock In \emph{Proceedings of the IEEE/CVF international conference on computer vision}, pages 774--782, 2021.

\bibitem[Netzer et~al.(2011)Netzer, Wang, Coates, Bissacco, Wu, and Ng]{netzer2011reading}
Yuval Netzer, Tao Wang, Adam Coates, Alessandro Bissacco, Bo Wu, and Andrew~Y Ng.
\newblock Reading digits in natural images with unsupervised feature learning.
\newblock 2011.

\bibitem[Oquab et~al.(2023)Oquab, Darcet, Moutakanni, Vo, Szafraniec, Khalidov, Fernandez, Haziza, Massa, El-Nouby, et~al.]{oquab2023dinov2}
Maxime Oquab, Timoth{\'e}e Darcet, Th{\'e}o Moutakanni, Huy Vo, Marc Szafraniec, Vasil Khalidov, Pierre Fernandez, Daniel Haziza, Francisco Massa, Alaaeldin El-Nouby, et~al.
\newblock Dinov2: Learning robust visual features without supervision.
\newblock \emph{arXiv preprint arXiv:2304.07193}, 2023.

\bibitem[Peng et~al.(2017)Peng, Zhang, Yu, Luo, and Sun]{peng2017large}
Chao Peng, Xiangyu Zhang, Gang Yu, Guiming Luo, and Jian Sun.
\newblock Large kernel matters--improve semantic segmentation by global convolutional network.
\newblock In \emph{Proceedings of the IEEE conference on computer vision and pattern recognition}, pages 4353--4361, 2017.

\bibitem[Richter et~al.(2021)Richter, J, A., and U]{Richter_2021}
Mats~L. Richter, Sch\"oning J, Wiedenroth A., and Krumnack U.
\newblock Should you go deeper? optimizing convolutional neural network architectures without training.
\newblock \emph{arXiv preprint arXiv:2106.12307v2}, 2021.

\bibitem[Shen et~al.(2017)Shen, Gan, Yan, and Zeng]{shen2017semantic}
Falong Shen, Rui Gan, Shuicheng Yan, and Gang Zeng.
\newblock Semantic segmentation via structured patch prediction, context crf and guidance crf.
\newblock In \emph{Proceedings of the IEEE Conference on Computer Vision and Pattern Recognition}, pages 1953--1961, 2017.

\bibitem[Simonyan and Zisserman(2014)]{simonyan2014very}
Karen Simonyan and Andrew Zisserman.
\newblock Very deep convolutional networks for large-scale image recognition.
\newblock \emph{arXiv preprint arXiv:1409.1556}, 2014.

\bibitem[Sudhakaran et~al.(2020)Sudhakaran, Escalera, and Lanz]{Sudhakaran_2020}
Swathikiran Sudhakaran, Sergio Escalera, and Oswald Lanz.
\newblock Gate-shift networks for video action recognition.
\newblock In \emph{Proceedings of the IEEE/CVF Conference on Computer Vision and Pattern Recognition}, pages 1102--1111, 2020.

\bibitem[Sudhakaran et~al.(2023)Sudhakaran, Escalera, and Lanz]{sudhakaran2022gate}
Swathikiran Sudhakaran, Sergio Escalera, and Oswald Lanz.
\newblock Gate-shift-fuse for video action recognition.
\newblock \emph{IEEE Transactions on Pattern Analysis and Machine Intelligence}, 2023.

\bibitem[Sun et~al.(2016)Sun, Xie, and Pu]{sun2016mixed}
Haiming Sun, Di Xie, and Shiliang Pu.
\newblock Mixed context networks for semantic segmentation.
\newblock \emph{arXiv preprint arXiv:1610.05854}, 2016.

\bibitem[Szegedy et~al.(2015)Szegedy, Liu, Jia, Sermanet, Reed, Anguelov, Erhan, Vanhoucke, and Rabinovich]{szegedy2015going}
Christian Szegedy, Wei Liu, Yangqing Jia, Pierre Sermanet, Scott Reed, Dragomir Anguelov, Dumitru Erhan, Vincent Vanhoucke, and Andrew Rabinovich.
\newblock Going deeper with convolutions.
\newblock In \emph{Proceedings of the IEEE conference on computer vision and pattern recognition}, pages 1--9, 2015.

\bibitem[Szegedy et~al.(2016)Szegedy, Vanhoucke, Ioffe, Shlens, and Wojna]{szegedy2016rethinking}
Christian Szegedy, Vincent Vanhoucke, Sergey Ioffe, Jon Shlens, and Zbigniew Wojna.
\newblock Rethinking the inception architecture for computer vision.
\newblock In \emph{Proceedings of the IEEE conference on computer vision and pattern recognition}, pages 2818--2826, 2016.

\bibitem[Tan and Le(2019)]{tan2019efficientnet}
Mingxing Tan and Quoc Le.
\newblock Efficientnet: Rethinking model scaling for convolutional neural networks.
\newblock In \emph{International conference on machine learning}, pages 6105--6114. PMLR, 2019.

\bibitem[Tian et~al.(2020)Tian, Lin, Sun, Zhou, Yan, and Ouyang]{tian2020improving}
Keyu Tian, Chen Lin, Ming Sun, Luping Zhou, Junjie Yan, and Wanli Ouyang.
\newblock Improving auto-augment via augmentation-wise weight sharing.
\newblock \emph{Advances in Neural Information Processing Systems}, 33:\penalty0 19088--19098, 2020.

\bibitem[Touvron et~al.(2021)Touvron, Cord, Sablayrolles, Synnaeve, and J{\'e}gou]{touvron2021going}
Hugo Touvron, Matthieu Cord, Alexandre Sablayrolles, Gabriel Synnaeve, and Herv{\'e} J{\'e}gou.
\newblock Going deeper with image transformers.
\newblock In \emph{Proceedings of the IEEE/CVF international conference on computer vision}, pages 32--42, 2021.

\bibitem[van Amersfoort et~al.(2017)van Amersfoort, Kannan, Ranzato, Szlam, Tran, and Chintala]{Joost_2017}
Joost~R. van Amersfoort, A. Kannan, Marc'Aurelio Ranzato, Arthur Szlam, Du Tran, and Soumith Chintala.
\newblock Transformation-based models of video sequences.
\newblock \emph{ArXiv}, abs/1701.08435, 2017.

\bibitem[Wang et~al.(2017)Wang, Jiang, Qian, Yang, Li, Zhang, Wang, and Tang]{wang2017residual}
Fei Wang, Mengqing Jiang, Chen Qian, Shuo Yang, Cheng Li, Honggang Zhang, Xiaogang Wang, and Xiaoou Tang.
\newblock Residual attention network for image classification.
\newblock In \emph{Proceedings of the IEEE conference on computer vision and pattern recognition}, pages 3156--3164, 2017.

\bibitem[Wang et~al.(2016)Wang, Xiong, Wang, Qiao, Lin, Tang, and Van~Gool]{Wang_2016}
Limin Wang, Yuanjun Xiong, Zhe Wang, Yu Qiao, Dahua Lin, Xiaoou Tang, and Luc Van~Gool.
\newblock Temporal segment networks: Towards good practices for deep action recognition.
\newblock In \emph{European conference on computer vision}, pages 20--36. Springer, 2016.

\bibitem[Wang et~al.(2018{\natexlab{a}})Wang, Chen, Yuan, Liu, Huang, Hou, and Cottrell]{wang2018understanding}
Panqu Wang, Pengfei Chen, Ye Yuan, Ding Liu, Zehua Huang, Xiaodi Hou, and Garrison Cottrell.
\newblock Understanding convolution for semantic segmentation.
\newblock In \emph{2018 IEEE winter conference on applications of computer vision (WACV)}, pages 1451--1460. Ieee, 2018{\natexlab{a}}.

\bibitem[Wang et~al.(2021)Wang, Cheng, Chen, Tang, and Hsieh]{wang2021rethinking}
Ruochen Wang, Minhao Cheng, Xiangning Chen, Xiaocheng Tang, and Cho-Jui Hsieh.
\newblock Rethinking architecture selection in differentiable nas.
\newblock In \emph{International Conference on Learning Representation}, 2021.

\bibitem[Wang et~al.()Wang, Dai, Chen, Huang, Li, Zhu, Hu, Lu, Lu, Li, et~al.]{wang2211internimage}
W Wang, J Dai, Z Chen, Z Huang, Z Li, X Zhu, X Hu, T Lu, L Lu, H Li, et~al.
\newblock Internimage: Exploring large-scale vision foundation models with deformable convolutions. arxiv 2022.
\newblock \emph{arXiv preprint arXiv:2211.05778}.

\bibitem[Wang et~al.(2018{\natexlab{b}})Wang, Girshick, Gupta, and He]{wang2018non}
Xiaolong Wang, Ross Girshick, Abhinav Gupta, and Kaiming He.
\newblock Non-local neural networks.
\newblock In \emph{Proceedings of the IEEE conference on computer vision and pattern recognition}, pages 7794--7803, 2018{\natexlab{b}}.

\bibitem[Wortsman et~al.(2022)Wortsman, Ilharco, Gadre, Roelofs, Gontijo-Lopes, Morcos, Namkoong, Farhadi, Carmon, Kornblith, et~al.]{wortsman2022model}
Mitchell Wortsman, Gabriel Ilharco, Samir~Ya Gadre, Rebecca Roelofs, Raphael Gontijo-Lopes, Ari~S Morcos, Hongseok Namkoong, Ali Farhadi, Yair Carmon, Simon Kornblith, et~al.
\newblock Model soups: averaging weights of multiple fine-tuned models improves accuracy without increasing inference time.
\newblock In \emph{International Conference on Machine Learning}, pages 23965--23998. PMLR, 2022.

\bibitem[Wu et~al.(2016)Wu, Shen, and Hengel]{wu2016bridging}
Zifeng Wu, Chunhua Shen, and Anton van~den Hengel.
\newblock Bridging category-level and instance-level semantic image segmentation.
\newblock \emph{arXiv preprint arXiv:1605.06885}, 2016.

\bibitem[Wu et~al.(2019)Wu, Shen, and Van Den~Hengel]{wu2019wider}
Zifeng Wu, Chunhua Shen, and Anton Van Den~Hengel.
\newblock Wider or deeper: Revisiting the resnet model for visual recognition.
\newblock \emph{Pattern Recognition}, 90:\penalty0 119--133, 2019.

\bibitem[Xie et~al.(2017)Xie, Girshick, Doll{\'a}r, Tu, and He]{xie2017aggregated}
Saining Xie, Ross Girshick, Piotr Doll{\'a}r, Zhuowen Tu, and Kaiming He.
\newblock Aggregated residual transformations for deep neural networks.
\newblock In \emph{Proceedings of the IEEE conference on computer vision and pattern recognition}, pages 1492--1500, 2017.

\bibitem[Zhang et~al.(2022)Zhang, Wu, Zhang, Zhu, Lin, Zhang, Sun, He, Mueller, Manmatha, Li, and Smola]{Zhang_2022}
Hang Zhang, Chongruo Wu, Zhongyue Zhang, Yi Zhu, Haibin Lin, Zhi Zhang, Yue Sun, Tong He, Jonas Mueller, R. Manmatha, Mu Li, and Alexander Smola.
\newblock Resnest: Split-attention networks.
\newblock In \emph{2022 IEEE/CVF Conference on Computer Vision and Pattern Recognition Workshops (CVPRW)}, pages 2735--2745, 2022.

\bibitem[Zhao et~al.(2017)Zhao, Shi, Qi, Wang, and Jia]{zhao2017pyramid}
Hengshuang Zhao, Jianping Shi, Xiaojuan Qi, Xiaogang Wang, and Jiaya Jia.
\newblock Pyramid scene parsing network.
\newblock In \emph{Proceedings of the IEEE conference on computer vision and pattern recognition}, pages 2881--2890, 2017.

\bibitem[Zheng et~al.(2021)Zheng, Lu, Zhao, Zhu, Luo, Wang, Fu, Feng, Xiang, Torr, et~al.]{zheng2021rethinking}
Sixiao Zheng, Jiachen Lu, Hengshuang Zhao, Xiatian Zhu, Zekun Luo, Yabiao Wang, Yanwei Fu, Jianfeng Feng, Tao Xiang, Philip~HS Torr, et~al.
\newblock Rethinking semantic segmentation from a sequence-to-sequence perspective with transformers.
\newblock In \emph{Proceedings of the IEEE/CVF conference on computer vision and pattern recognition}, pages 6881--6890, 2021.

\bibitem[Zoph and Le(2016)]{zoph2016neural}
Barret Zoph and Quoc~V Le.
\newblock Neural architecture search with reinforcement learning.
\newblock \emph{arXiv preprint arXiv:1611.01578}, 2016.

\bibitem[Zoph et~al.(2018)Zoph, Vasudevan, Shlens, and Le]{zoph2018learning}
Barret Zoph, Vijay Vasudevan, Jonathon Shlens, and Quoc~V Le.
\newblock Learning transferable architectures for scalable image recognition.
\newblock In \emph{Proceedings of the IEEE conference on computer vision and pattern recognition}, pages 8697--8710, 2018.

\end{thebibliography}


\begin{thebibliography}{9}
    \bibitem{reference1}
    Ekin D Cubuk, Barret Zoph, Dandelion Mane, Vijay Vasudevan, and Quoc V Le. Autoaugment: Learning augmentation strategies from data. In \emph{Proceedings of the IEEE/CVF conference on computer vision and pattern recognition}, pages 113–123, 2019.
    
    \bibitem{reference2}
    Ekin D Cubuk, Barret Zoph, Jonathon Shlens, and Quoc V Le. Randaugment: Practical automated data augmentation with a reduced search space. In \emph{Proceedings of the IEEE/CVF conference on computer vision and pattern recognition workshops}, pages 702--703, 2020.
\end{thebibliography}
